\documentclass{clv2}
\issue{-}{-}{-} 
\usepackage{amsfonts}
\usepackage[inline]{enumitem}
\usepackage{graphicx, subfigure}
\usepackage{listings}
\usepackage{booktabs,amsmath}
\usepackage{rotating}
\usepackage{multirow}
\usepackage{changepage}

\newcommand{\argmax}[1]{\underset{#1}{\operatorname{arg}\,\operatorname{max}}\;}

\runningtitle{A Game-Theoretic Approach to WSD}
\runningauthor{R. Tripodi, M. Pelillo}

\historydates{Submission received: 8th June, 2015; Revised submission received:
25th January, 2016;Accepted for publication: 24th March, 2016; will appear in
either the 42-4 issue (December 2016) or the 43-1 issue (March 2017) }

\begin{document}

\title{A Game-Theoretic Approach to Word Sense Disambiguation}
\author{Rocco Tripodi \thanks{European Centre for Living Technology, Ca' Minich, S. Marco 2940 30124 Venezia, Italy. E-mail: rocco.tripodi@unive.it}}
\affil{Ca' Foscari University of Venice}

\author{Marcello Pelillo$^{*,}$\thanks{Dipartimento di Scienze Ambientali, Informatica e Statistica, Via Torino 155 - 30172 Venezia, Italy. E-mail: pelillo@unive.it}}
\affil{Ca' Foscari University of Venice}
\maketitle

\begin{abstract} This paper presents a new model for word sense disambiguation
formulated in terms of evolutionary game theory, where each word to be
disambiguated is represented as a node on a graph whose edges represent word
relations and senses are represented as classes. The words simultaneously update
their class membership preferences according to the senses that neighboring
words are likely to choose. We use distributional information to weigh the
influence that each word has on the decisions of the others and semantic
similarity information to measure the strength of compatibility among the
choices. With this information we can formulate the word sense disambiguation
problem as a constraint satisfaction problem and solve it using tools derived
from game theory, maintaining the textual coherence. The model is based on two
ideas: similar words should be assigned to similar classes and the meaning of a
word does not depend on all the words in a text but just on some of them. The
paper provides an in-depth motivation of the idea of modeling the word sense
disambiguation problem in terms of game theory, which is illustrated by an
example. The conclusion presents an extensive analysis on the combination of
similarity measures to use in the framework and a comparison with
state-of-the-art systems. The results show that our model outperforms
state-of-the-art algorithms and can be applied to different tasks and in
different scenarios. \end{abstract}

\section{Introduction} Word Sense Disambiguation (WSD) is the task of
identifying the intended meaning of a word based on the context in which it
appears \cite{navigli2009word}. It has been studied since the beginnings of
Natural Language Processing (NLP) \cite{weaver1955translation} and today it is
still a central topic of this discipline. This because it is important for many
NLP tasks such as text understanding \cite{kilgarriff1997don}, text entailment
\cite{dagan2004probabilistic}, machine translation \cite{vickrey2005word},
opinion mining \cite{smrvz2006using}, sentiment analysis
\cite{rentoumi2009sentiment} and information extraction \cite{zhong2012word}.
All these applications can benefit from the disambiguation of ambiguous words,
as a preliminary process; otherwise they remain on the surface of the word,
compromising the coherence of the data to be analyzed
\cite{pantel2002discovering}. 

 To solve this problem, over the past few
years, the research community has proposed several algorithms during the years,
based on supervised ~\cite{zhong2010makes,tratz2007pnnl}, semi-supervised
\cite{pham2005word,navigli2005structural} and unsupervised
~\cite{mihalcea2005unsupervised,mccarthy2007unsupervised} learning models.
Nowadays, although supervised methods perform better in general domains,
unsupervised and semi-supervised models are receiving increasing attention from
the research community with performances close to the state of the art of
supervised systems ~\cite{ponzetto2010knowledge}. In particular knowledge-based
and graph-based algorithms are emerging as promising approaches to solve the
problem ~\cite{agirre2009knowledge,sinha2007unsupervised}. The peculiarities of
these algorithms are that they do not require any corpus evidence and use only
the structural properties of a lexical database to perform the disambiguation
task. In fact, unsupervised methods are able to overcome a common problem in
supervised learning: the knowledge acquisition problem, which consists in the
production of large-scale resources, manually annotated with word senses.

Knowledge-based approaches exploit the information from knowledge resources such
as dictionaries, thesauri or ontologies and compute sense similarity scores to
disambiguate words in context \cite{mihalcea2006knowledge}. Graph-based
approaches model the relations among words and senses in a text with graphs,
representing words and senses as nodes and the relations among them as edges.
From this representation the structural properties of the graph can be extracted
and the most relevant concepts in the network can be computeds
\cite{navigli2007graph,agirre2006two}. 

 Our approach falls into these two
lines of research; it uses a graph structure to model the \emph{geometry} of the
data points (the words in a text) and a knowledge base to extract the senses of
each word and to compute the similarity among them. The most important
difference between our approach and state-of-the-art graph-based approaches
\cite{moro2014entity,agirre2014random,navigli2010experimental,sinha2007unsupervised,veronis2004hyperlex}
is that in our method the graph contains only words and not senses. This graph
is used to model the pairwise interaction among words and not to rank the senses
in the graph according to their relative importance.


The starting point of our research is
based on two fundamental assumptions: \begin{enumerate}
\item the meaning of a sentence emerges from the interaction of the components
      which are involved in it;
\item these interactions are different and must be weighted in order to supply
      the right amount of information. \end{enumerate} \noindent We interpret
      language as a complex adaptive system, composed of linguistic units and
      their interactions \cite{cong2014approaching,larsen2008complex}. The
      interactions among units give rise to the emergence of properties, which in
      our case, by problem definition, can be interpreted as meanings. In our
      model the relations between the words are weighted by a similarity measure
      with a distributional approach, increasing the weights among words which
      share a proximity relation. Weighting the interaction of
      the nodes in the graph is helpful in situations in which the
      indiscriminate use of contextual information can deceive. Furthermore, it
      models the idea that the meaning of a word does not depend on all the
      words in a text but just on some of them \cite{chaplot2015unsupervised}.

      This problem is illustrated in the sentences below:
\begin{itemize}
  \item There is a financial institution near the river bank.
  \item They were troubled by insects while playing cricket. \end{itemize}
        \noindent In these two sentences\footnote{A complete example of the
        disambiguation of the first sentence is given in Section
        \ref{sec:esempio}} the meaning of the words \emph{bank} and
        \emph{cricket} can be misinterpreted by a centrality algorithm that
        tries to find the most important node in the graph composed of all the
        possible senses of the words in the sentence. This because the meanings
        of the words \emph{financial} and \emph{institution} tend to shift the
        meaning of the word \emph{bank} toward its financial meaning and not
        toward its naturalistic meaning. The same behavior can be observed for
        the word \emph{cricket}, which is shifted by the word \emph{insect}
        toward its insect meaning and not toward its game meaning. In our work
        the disambiguation task is performed imposing a stronger importance on
        the relations between the words \emph{bank} and \emph{river} for the
        first sentence and between \emph{cricket} and \emph{play} for the
        second; exploiting proximity relations.

 Our approach is based on the
        principle that the senses of the words that share a strong relation must
        be similar. The idea of assigning a similar class to similar objects has
        been implemented in a different way by Kleinberg and Tardos
        \shortcite{kleinberg2002approximation}, within a Markow random field
        framework. They have shown that it is beneficial in combinatorial
        optimization problems. In our case, this idea can preserve the
        textual coherence; a characteristic that is missing in many
        state-of-the-art systems. In particular, it is missing in systems in
        which the words are disambiguated independently. On the contrary, our
        approach disambiguates all the words in a text concurrently, using an
        underlying structure of interconnected links, which models the
        interdependence between the words. In so doing, we model the idea that
        the meaning for any word depends at least implicitly on the combined
        meaning of all the interacting words. 

 In our study, we model these
        interactions by developing a system in which it is possible to map
        lexical items onto concepts exploiting contextual information in a way
        in which collocated words influence each other simultaneously, imposing
        constraints in order to preserve the textual coherence. For this reason,
        we have decided to use a powerful tool, derived from game theory: the
        non-cooperative games (see Section \ref{sec:GT}). In our system, the
        nodes of the graph are interpreted as players, in the game theoretic
        sense (see Section \ref{sec:GT}), which play a game with the other words
        in the graph, in order to maximize their utility; constraints are
        defined as similarity measures among the senses of two words that are
        playing a game. The concept of utility has been used in different ways
        in the game theory literature, in general, it refers to the satisfaction
        that a player derives from the outcome of a game
        \cite{szabo2007evolutionary}. From our point of view, increasing the
        utility of a word means increasing the textual coherence, in a
        distributional semantics perspective \cite{firth1975synopsis}. In fact,
        it has been shown that collocated words tend to have a determined
        meaning
\cite{gale1992method,yarowsky1993one}. 

 Game theoretic frameworks have been
used in different ways to study the language use
~\cite{pietarinen2007game,skyrms2010signals} and evolution
~\cite{nowak2001evolution}, but to the best of our knowledge, our method is the
first attempt to use it in a specific NLP task. This choice is motivated by the
fact that game theoretic models are able to perform a consistent labeling of the
data \cite{hummel1983foundations,pelillo1997dynamics}, taking into account
contextual information. These features are of great importance for an
unsupervised or semi-supervised algorithm, which tries to perform a WSD task,
because, by assumption, the sense of a word is given by the context in which it
appears. Within a game theoretic framework we are able to cast the WSD problem
as a continuous optimization problem, exploiting contextual information in a
dynamic way. Furthermore, no supervision is required and the system can adapt
easily to different contextual domains, which is exactly what is required for a
WSD algorithm.

 The additional reason for the use of a consistent labeling
system relies on the fact that it is able to deal with \emph{semantic drifts}
\cite{curran2007minimising}. In fact, as shown in the above two sentences,
concentrating the disambiguation task of a word on highly collocated words,
taking into account proximity (or even syntactic) information allows the meaning
interpretation to be guided only towards senses which are strongly related to the
word which has to be disambiguated. 

 In this
article, we provide a detailed discussion about the motivation behind our
approach and a full evaluation of our algorithm comparing it with
state-of-the-art systems, in WSD tasks. In a previous work we used a similar algorithm in
a semi-supervised scenario \cite{tripodi2015nlpcs}, casting the WSD task as a graph
transduction problem. Now we have extended that work making the algorithm fully
unsupervised. Furthermore, in this article we provide a complete evaluation of
the algorithm extending our previous works \cite{tripodi2015semeval}, exploiting proximity relations among words.

An important feature of our approach is that it is versatile. In fact, the
method can adapt to different scenarios and to different tasks and it is
possible to use it as unsupervised or semi-supervised. The semi-supervised
approach, presented in \cite{tripodi2015nlpcs}, is a bootstrapping graph based
method, which propagates, over the graph, the information from labeled nodes to
unlabeled. In this article, we also provide a new semi-supervised version of the
approach, which can exploit the evidence from sense tagged corpora or the most
frequent sense heuristic and does not require labeled nodes to propagate the
labeling information. 

 We tested our approach on different datasets, from
WSD and entity linking tasks, in order to find the similarity measures, which
perform better and evaluated it against unsupervised, semi-supervised and
supervised state-of-the-art systems. The results of this evaluation shows that
our method performs well and can be considered as a valid alternative to current
models.


\section{Related Works}\label{sec:relWork} There are two major paradigms in WSD:
supervised and knowledge-based. Supervised algorithms learn, from sense-labeled
corpora, a computational model of the words of interest. Then, the obtained
model is used to classify new instances of the same words. Knowledge-based
algorithms perform the disambiguation task by using an existing lexical
knowledge base, which usually is structured as a semantic network. Then, these
approaches use graph algorithms to disambiguate the words of interests, based on
the relations that these words' senses have in the network
\cite{pilehvar2014large}.

 A popular supervised WSD system, which has shown
good performances in different WSD tasks, is \emph{It Makes Sense} (IMS)
\cite{zhong2010makes}. It takes as input a text and for each content word (noun,
verb, adjective, or adverb) outputs a list of possible senses ranked according
to the likelihood of appearing in a determined context and extracted from a
knowledge base. The training data used by this system are derived from SemCor
\cite{miller1993semantic}, DSO \cite{ng1996integrating} and collected
automatically exploiting parallel corpora \cite{chan2005scaling}. Its default
classifier is LIBLINEAR\footnote{http://liblinear.bwaldvogel.de} with a linear
kernel and its default parameters.

 Unsupervised and knowledge-based
algorithms for WSD are attracting great attention from the research community.
This because, supervised systems require training data, which are difficult to
obtain. In fact, producing sense tagged data is a time-consuming process, which
has to be carried out separately for each language of interest. Furthermore, as
investigated by Yarowsky and Florian \shortcite{yarowsky2002evaluating}, the
performances of a supervised algorithm degrade substantially with the increasing
of sense entropy. Sense entropy refers to the distribution over the possible
senses of a word, as seen in training data. Additionally, a supervised system
has problems to adapt to different contexts, because it depends on prior
knowledge, which makes the algorithm rigid, therefore can not efficiently adapt
to domain specific cases, when other optimal solution may be available
\cite{yarowsky2002evaluating}.

 One of the most common heuristics that allows
to exploit sense tagged data such as SemCor \cite{miller1993semantic} is the
most frequent sense. It exploits the overall sense distribution for each word to
be disambiguated, choosing the sense with the highest probability regardless of
any other information. This simple procedure is very powerful in general domains
but can not handle senses with a low distribution, which could be found in
specific domains. 

 With these observations in mind Koeling et al.
\shortcite{koeling2005domain} created three domain specific corpora to evaluate
WSD systems. They tested whether WSD algorithms are able to adapt to different
contexts, comparing their results with the most frequent sense heuristic,
computed on general domains corpora. They used an unsupervised approach to
obtain the most frequent sense for a specific domain
\cite{mccarthy2007unsupervised} and demonstrated that their approach outperforms
the most frequent sense heuristic derived from general domain and labeled
data.

 This heuristics, for the unsupervised acquisition of the predominant
sense of a word, consists in collecting all the possible senses of a word and
then in ranking these senses. The ranking is computed according to the
information derived from a distributional thesaurus automatically produced from
a large corpus and a semantic similarity measure derived from the sense
inventory. Although the authors have demonstrated that this approach is able to
outperform the most frequent sense heuristic computed on sense tagged data on
general domains, it is not easy to use it on real world applications, especially
when the domain of the text to be disambiguated is not known in advance.

Other unsupervised and semi-supervised approaches, instead of computing the
prevalent sense of a word, try to identify the actual sense of a word in a
determined phrase, exploiting the information derived from its context. This is
the case of traditional algorithms, which exploit the pairwise semantic
similarity among a target word and the words in its context
\cite{lesk1986automatic,resnik1995using,patwardhan2003using}. Our work could be
considered as a continuation of this tradition, which tries to identify the
intended meaning of a word given its context, using a new approach for the
computation of the sense combinations. 

 Graph-based algorithms for WSD are
gaining much attention in the NLP community. This is because graph theory is a
powerful tool that can be employed both for the organization of the contextual
information and for the computation of the relations among word senses. It
allows to extract the structural properties of a text. Examples of this kind of
approaches construct a graph from all the senses of the words in a text and then
use connectivity measures in order to identify the most relevant word senses in
the graph \cite{sinha2007unsupervised,navigli2007graph}.
Navigli and Lapata \shortcite{navigli2007graph} conducted an extensive analysis
of graph connectivity measures for unsupervised WSD. Their approach uses a
knowledge base, such as WordNet, to collect and organize all the possible senses
of the words to be disambiguated in a graph structure, then uses the same
resource to search for a path (of predefined length) between each pair of senses
in the graph and if it exists, it adds all the nodes and edges on this path to
the graph. These measures analyze local and global properties of the graph.
Local measures, such as degree centrality and eigenvector centrality, determine
the degree of relevance of a single vertex. Global properties, such as
compactness, graph entropy and edge density, analyze the structure of the graph
as a whole. The results of the study show that local measures outperform global
measure and in particular, degree centrality and PageRank
\cite{page1999pagerank} (which is a variant of the eigenvector centrality
measure) achieve the best results.

 PageRank \cite{page1999pagerank} is one
of the most popular algorithms for WSD, in fact, it was implemented in different
ways by the research community
~\cite{mihalcea2004pagerank,haveliwala2002topic,agirre2014random,de2010robust}.
It represents the senses of the words in a text as nodes of a graph. It uses a
knowledge base to collect the senses of the words in a text and represents them
as nodes of a graph. The structure of this resource is used to connect each node
with its related senses in a directed graph. The main idea of this algorithm is
that whenever a link from a node to another exists, a vote is produced,
increasing the rank of the voted node. It works by counting the number and
quality of links to a node in order to determine an estimation of how important
the node is in the network. The underlying assumption is that more important
nodes are likely to receive more links from other nodes \cite{page1999pagerank}.
Exploiting this idea the ranking of the nodes in the graph can be computed
iteratively with the following equation:%
\begin{equation}\label{eq:pagerank}
  Pr = cMPr+(1-c)v
\end{equation} \noindent%
where $M$ is the transition matrix of the graph, $v$ is a $N \times 1$ vector
representing a probability distribution and $c$ is the so-called damping factor
that represents the chance that the process stops, restarting from a random
node. At the end of the process each word is associated with the most important
concept related to it. One problem of this framework is that the labeling
process is not assumed to be consistent. 

 An algorithm, which tries to
improve centrality algorithms, is SUDOKU, introduced by Minion and Sainudiin
\shortcite{manion2014iterative}. It is an iterative approach, which
simultaneously constructs the graph and disambiguates the words using a
centrality function. It starts inserting the nodes corresponding to the senses
of the words with low polysemy and and iteratively inserting the more ambiguous
words. The advantages of this method are that the use of small graphs, at the
beginning of the process, reduces the complexity of the problem and that it can
be used with different centrality measures. 

 Recently a new model for WSD
has been introduced, based on an undirected graphical model
\cite{chaplot2015unsupervised}. It approaches the WSD problem as a maximum a
posteriori query on a Markov random field \cite{jordan2002graphical}. The
graph is constructed using the content words of a sentence as nodes and
connecting them with edges if they share a relation, determined using a
dependency parser. The values that each node in the graphical model can take
include the senses of the corresponding word. The senses are collected using a
knowledge base and weighted using a probability distribution based on the
frequency of the senses in the knowledge base. Furthermore, the senses between
two related words are weighted using a similarity measure. The goal of this
approach is to maximize the joint probability of the senses of all the words in
the sentence, given dependency structure of the sentence, the frequency of the
senses and the similarity among them. 

 A new graph based, semi-supervised
approach, introduced to deal with multilingual WSD \cite{navigli2012joining} and
entity inking problems, is Babelfy \cite{moro2014entity}. Multilingual WSD is
an important task because traditional WSD algorithms and resources are focused
on English language. It exploits the information from large multilingual
knowledge, such as BabelNet \cite{navigli2012babelnet} to perform this task.
Entity linking consists in disambiguating the named entities in a text and in
finding the appropriate resources in an ontology, which correspond to the specific
entities mentioned in a text.
Babelfy creates the semantic signature of each word to be disambiguated, that
consists in collecting, from a semantic network, all the nodes related to a
particular concepts, exploiting the global structure of the network. This
process leads to the construction of a graph-based representation of the whole
text. Then, it applies Random Walk with Restart \cite{tong2006fast} to find the
most important nodes in the network, solving the WSD problem. 

 Approaches
which are more similar to ours in the formulation of the problem have been
described by Araujo \shortcite{araujo2007evolutionary}. The author reviewed the
literature devoted to the application of different evolutionary algorithm to
several aspects of NLP: syntactical analysis, grammar induction, machine
translation, text summarization, semantic analysis, document clustering and
classification. Basically these approaches are search and optimization methods
inspired by biological evolution principles. A specific evolutionary approach
for WSD has been introduced by Menai \shortcite{menai2014word}. It uses genetic
algorithms \cite{holland1975adaptation} and memetic algorithms
\cite{moscato1989evolution} in order to improve the performances of a
\emph{gloss-based} method. It assumes that there is a population of
individuals, represented by all the senses of the words to be disambiguated, and
that there is a selection process, which selects the best candidates in the
population. The selection process is defined as a sense similarity function,
which gives a higher score to candidates with specific features, increasing
their \emph{fitness} to the detriment of the other population members. This
process is repeated until the \emph{fitness} level of the population regularizes
and at the end the candidates with higher \emph{fitness} are selected as
solutions of the problem. Another approach, which address the disambiguation
problem in terms of space search is GETALP \cite{schwab2013getalp}, it uses an
Ant Colony algorithm to find the best path in the weighted graph constructed
measuring the similarity of all the senses in a text and assigning to each word
to be disambiguated the sense corresponding to the node in this path. 

 These
methods are similar to our study in the formulation of the problem; the main
difference is that our approach is defined in terms of evolutionary game theory.
As it is shown in the next section, this approach ensures that the final
labeling of the data is consistent and that the solution of the problem is
always found. In fact, our system always converges to the nearest Nash 
equilibrium from  where the dynamics have been started. 

\section{Word Sense Disambiguation as a Consistent Labeling Problem} WSD can be
interpreted as a sense-labeling task \cite{navigli2009word}, which consists in
assigning a sense label to a target word. As a labeling problem we need an
algorithm, which performs this task in a consistent way, taking into account the
context in which the target word occurs. Following this observation we can
formulate the WSD task as a constraint satisfaction problem
\cite{tsang1995foundations} in which the labeling process has to satisfy some
constraints in order to be consistent. This approach gives us the possibility
not only to exploit the contextual information of a word but also to find the
most appropriate sense association for the target word and the words in its
context. This is the most important contribution of our work, which
distinguishes it from existing WSD algorithms. In fact, in some cases using only
contextual information without the imposition of constraints can lead to
inconsistencies in the assignment of senses to related words. 

 As an
illustrative example we can consider a binary CSP, which is defined by a set of
variables representing the elements of the problem and a set of binary
constraints representing the relationships among variables. The problem is
considered solved if there is a solution, which satisfies all the constraints.
This setting can be described in a formal manner as a triple $(V,D,R)$, where $V
= \{v_1,...,v_n\}$ is the set of variables, $D = \{D_{v_1} ,..., D_{v_n} \}$ is
the set of domains for each variable, each $D_{v_i}$ denoting a finite set of
possible values for variable $v_i$; and $R = \{R_{ij} | R_{ij} \subseteq D_{v_i}
\times D_{v_j}\}$ is a set of binary constraints where $R_{ij}$ describe a set
of compatible pairs of values for the variables $v_i$ and $v_j$. $R$ can be
defined as a binary matrix of size $p \times q$ where $p$ and $q$ are the
cardinalities of domains and variables respectively. Each element of the binary
matrix $R_{ij}(\lambda, \lambda^{'})=1$ indicates if the assignment $v_i
=\lambda$ is compatible with the assignment $v_j = \lambda^\prime_{i}$. $R$ is used to
impose constraints on the labeling so that each label assignment is consistent.

 The binary case described above assumes that the constraints are completely
violated or completely respected, which is restrictive; it is more appropriate,
in many real-word cases, to have a weight, which expresses the level of
confidence about a particular assignment \cite{hummel1983foundations}. This
notion of consistency has been shown to be related to the Nash equilibrium
concept in game theory \cite{miller1991copositive}. We have adopted this method
to approach the WSD task in order to perform a consistent labeling of the data.
In our case, we can consider variables as words, labels as word senses and
compatibility coefficients as similarity values among two word senses. To
explain how the Nash equilibria are computed we need to introduce basic notions
of game theory in the following sections.

\section{Game Theory}\label{sec:GT} In this section, we briefly introduce the basic
concepts of classical game theory and evolutionary game theory that we used in our framework; for a more detailed
analysis of these topics the reader is referred to
~\cite{weibull1997evolutionary,leyton2008essentials,sandholm2010population}.
\subsection{Classical Game Theory}
Game theory provides predictive power in interactive decision situations. 
It has been introduced by Von Neumann and Morgenstern \shortcite{von1944theory}
in order to develop a mathematical framework able to model the essentials of
decision making in interactive situations. In its normal form representation
(which is the one we use in this article) it consists of a finite set of
players $I=\{1,..,n\}$, a set of pure strategies for each player $S_i=\{s_1, ...,
s_m\}$, and a utility function $u_i : S_1 \times ... \times S_n \rightarrow
\mathbb{R}$, which associates strategies to payoffs. Each player can adopt a
strategy in order to play a game and the utility function depends on the
combination of strategies played at the same time by the players involved in the
game, not just on the strategy chosen by a single player. An important
assumption in game theory is that the players are rational and try to maximize
the value of $u_i$; Furthermore, in \emph{non-cooperative games} the players
choose their strategies independently, considering what the other players can
play and try to find the best strategy profile to employ in a game.

 A
strategy $s_i^*$ is said to be \emph{dominant} if and only if:
\begin{equation}\label{eq:domstrat}
  u_i(s_i^*,s_{-i})>u_i(s_i,s_{-i}), \forall s_{-i} \in S_{-i}
\end{equation}
\noindent where $S_{-i}$ represents all strategy sets other than player $i$'s.

 As an example, we can consider the famous \emph{Prisoner's Dilemma},
whose payoff matrix is shown in Table \ref{tab:prisoner}. Each cell of the
matrix represents a strategy profile, where the first number represents the
payoff of \emph{Player 1} ($P_1$) and the second is the payoff of \emph{Player
2} ($P_2$), when both players employ the strategy associated with a specific
cell. $P_1$ is called the \emph{row player} because it selects its strategy
according to the rows of the payoff matrix, $P_2$ is called the \emph{column
player} because it selects its strategy according to the columns of the payoff
matrix. In this game the strategy \emph{confess} is a \emph{dominant strategy}
for both players and this strategy combination is the \emph{Nash equilibrium} of
the game.

Nash equilibria represent the key concept of game theory and can be defined as
those strategy profiles in which each strategy is a best response to the
strategy of the co-player and no player has the incentive to unilaterally
deviate from his decision, because there is no way to do better. 

 In many
games, the players can also play \emph{mixed strategies}, which are probability
distributions over their pure strategies. Within this setting, the players
choose a strategy with a certain pre-assigned probability. A mixed strategy
set can be defined as a vector $x=(x_{1},\ldots,x_{m} )$, where $m$ is
the number of pure strategies and each component $x_{h}$ denotes the
probability that player $i$ chooses its $h$th pure strategy. For each player
its strategy set is defined as a standard simplex:
\begin{equation}\label{eq:simplex}
  \Delta = \Big\{ x \in \mathbb{R}^n :
\sum_{h=1}^m x_{h} = 1,\text{ and } x_{h} \geq 0 \text{ for all } h \in x \Big\}
\end{equation}
\noindent 
Each mixed strategy corresponds to a point on the simplex and its corners correspond to pure strategies.

 In a
\emph{two-players game} we can define a strategy profile as a pair $(p,q)$
where $p \in \Delta_i$ and $q \in \Delta_j$. The expected payoff for this
strategy profile is computed as follows: $u_i(p,q)=p \cdot A_i q$ and $u_j(p,q)=q
\cdot A_j p$, where $A_i$ and $A_j$ are the payoff matrices of player $i$ and player $j$, respectively.
The Nash equilibrium is computed in mixed strategies in the same way of pure
strategies. It is represented by a pair of strategies such that each is a best
response to the other. The only difference is that, in this setting, the
strategies are probabilities and must be computed considering the payoff matrix
of each player. 

 A game theoretic framework can be considered as a solid
tool in decision making situations since a fundamental theorem by Nash
\shortcite{nash1951non} states that any normal-form game has at least one mixed
Nash equilibrium, which can be employed as the solution of the decision problem.
\begin{table} \begin{center} \begin{tabular}{ l | c c } $P_1 \backslash P_2$ &
confess & don't confess \\ \hline confess & -5,-5 & 0,-6 \\ don't confess & -6,0
& -1,-1 \end{tabular} \end{center} \caption{\label{tab:prisoner} The Prisoner's
dilemma. } \end{table}

\subsection{Evolutionary Game Theory}\label{sec:EGT} Evolutionary game theory
has been introduced by Smith and Price \shortcite{smith1973conflict} overcoming
some limitations of traditional game theory, such as the hyper-rationality
imposed on the players. In fact, in real life situations the players choose a
strategy according to heuristics or social norms \cite{szabo2007evolutionary}.
It has been introduced in biology to explain the evolution of species. In this
context, strategies correspond to phenotypes (traits or behaviors), payoffs
correspond to offsprings, allowing players with a high actual payoff (obtained
thanks to their phenotype) to be more prevalent in the population. This
formulation explains natural selection choices among alternative phenotypes
based on their utility function. This aspect can be linked to rational choice
theory, in which players make a choice that maximizes their utility, balancing
cost against benefits \cite{okasha2012evolution}. 

 This intuition introduces
an \emph{inductive learning} process, in which we have a population of agents
which play games repeatedly with their neighbors. The players, at each
iteration, update their beliefs on the state of the game and choose their
strategy according to what has been effective and what has not in previous
games. The strategy space of each player $i$ is defined as a mixed strategy
profile $x_i$, as defined in the previous section, which lives in the mixed
strategy space of the game, given by the Cartesian product: \begin{equation}
\Theta = \times_{i \in I} \Delta_i \end{equation} The expected payoff of a pure
strategy $e^h$ in a single game is calculated as in mixed strategies. The
difference in evolutionary game theory is that a player can play the game with
all other players, obtaining a final payoff, which is the sum of all the partial
payoff obtained during the single games. We have that the payoff relatives to a
single strategy is: $u_i(e_i^h) = \sum_{j=1}^n(A_{ij} x_j)_h$ and the average
payoff $u_i(x) =\sum_{j=1}^n x_i^T A_{i j}x_j$, where $n$ is the number of
players with whom the games are played and $A_{ij}$ is the payoff matrix between
player $i$ and $j$. Another important characteristic of evolutionary game theory
is that the games are played repeatedly. In fact, at each iteration a player can
update its strategy space according to the payoffs gained during the games. He
can allocate more probability to the strategies with high payoff until an
equilibrium is reached. In order to find those states that correspond to the
Nash equilibria of the games, the replicator dynamic equation is used
\cite{taylor1978evolutionary}: \begin{equation}\label{eq:replicator}
\dot{x}=[u(e^h,x)-u(x,x)] \cdot x^h \text{ } \forall h \in x \end{equation}
\noindent which allows better than average strategies (best replies) to grow at
each iteration.

 The following theorem states that with equation
\ref{eq:replicator} it is always possible to find the Nash equilibria of the
games (see \cite{weibull1997evolutionary} for the proof).

\begin{theorem} A point $x \in \Theta$ is the limit of a trajectory of equation
\ref{eq:replicator} starting from the interior of $\Theta$ if and only if $x$ is
a Nash equilibrium. Further, if point $x \in \Theta$ is a strict Nash
equilibrium, then it is asymptotically stable, additionally implying that the
trajectories starting from all nearby states converge to $x$.\end{theorem}

As in \cite{erdem2012graph} we used the discrete time version of the replicator
dynamic equation for the experiments of this paper:
\begin{equation}\label{eq:replicatorDS} x^h(t+1)=x^h(t)\frac{u(e^h,x)}{u(x,x)}
\text{ } \forall h \in S \end{equation}\noindent where, at each time step $t$,
the players update their strategies according to the strategic environment,
until the system converges and the Nash equilibria are met. In classical
evolutionary game theory these dynamics describe a stochastic evolutionary
process in which the agents adapt their behaviors to the environment.

\begin{figure} \centering
\includegraphics[scale=0.25]{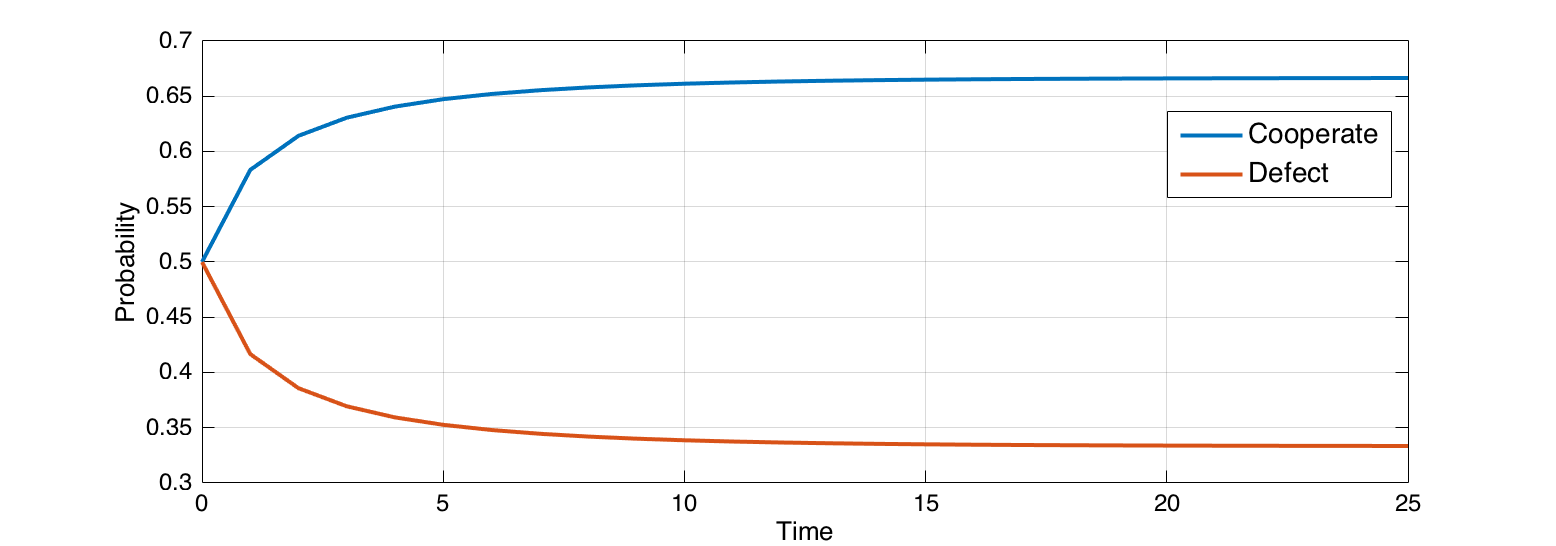} \caption{The dynamics of the
repeated prisoner's dilemma.}\label{fig:egtPris} \end{figure}

For example, if we analyze the prisoner's dilemma within the evolutionary game
theory framework we can see that the cooperative strategy (\emph{do not
confess}) tends to emerge as an equilibrium of the game and this is the best
situation for both players, because this strategy gives an higher payoff than
the defect strategy (\emph{confess}), which is the equilibrium in the classical
game theory framework. In fact, if the players play the game shown in Table
\ref{tab:prisoner} repeatedly and randomize their decisions in each game,
assigning at the beginning a normal distribution to their strategies, their
payoffs $u(x_{pi})$ can be computed as follows: \begin{itemize}
\item[] $u(x_{p1})=A_{p1}x_{p_2}=\begin{pmatrix} -5, & 0 \\ -6, & -1 \end{pmatrix}
 \begin{pmatrix} 0.5 \\ 0.5\end{pmatrix} = \begin{pmatrix} -2.5 \\ -3.5\end{pmatrix} $ 
 \item[] $u(x_{p2})=A_{p2}^Tx_{p_1}=\begin{pmatrix} -5, & -6 \\ 0, & -1
         \end{pmatrix}^T \begin{pmatrix} 0.5 \\ 0.5 \end{pmatrix} =
         \begin{pmatrix} -2.5 \\ -3.5\end{pmatrix} $  
 
 \end{itemize}

\noindent where $T$ is the transpose operator, required for $P_2$, which
chooses its strategies according to the columns of the matrix in Table
\ref{fig:egtPris}. This operation makes the matrices $A_{p1}$ and $A_{p2}$
identical and for this reason in this case the distinction among the two players
is not required since they get the same payoffs. Now we can compute the strategy
space of a player at time $t+1$ according to equation (\ref{eq:replicator}):
\begin{itemize}
\item[] $x_1$: $ -1.25 / -3 = 0.42 $
\item[] $x_2$: $ -1.75 / -3 = 0.58 $ \end{itemize}
The game is played with the new strategy spaces until the system converges, that
is when the difference among the payoffs at time $t_n$ and $t_{n-1}$ is under a
small threshold. In Figure \ref{fig:egtPris} we can see how the \emph{cooperate
strategy} increases over time, reaching a stationary point, which corresponds to
the equilibrium of the game. 

%

 Pietarinen \shortcite{pietarinen2007invitation} distinguishes two levels in

 The use of game theory as a tool to explain the origin of language ...

 The use of game theory as a tool to study communication systems relies on the

 In our work we interpret

\section{WSD Games}\label{sec:wsd-g} In this section we describe how the WSD
games are formulated. We assume that each player $i \in I$, which participates
in the games is a particular word in a text and that each strategy is a
particular word sense. The players can choose a determined strategy among the
set of strategies $S_i = \{1, ..., c\}$, each expressing a certain hypothesis
about its membership in a class and $c$ being the total number of classes
available. We consider $S_i$ as the mixed strategy for player $i$ as described
in Section \ref{sec:GT}. The games are played between two similar words, $i$ and
$j$, imposing only pairwise interaction between them. The payoff matrix $Z_{ij}$
of a single game is defined as a sense similarity matrix between the senses of
word $i$ and word $j$. The payoff function for each word is additively separable
and is computed as described in Section \ref{sec:EGT}.

 Formulating the
problem in this way we can apply equation (\ref{eq:replicatorDS}) to compute the
equilibrium state of the system, which corresponds to a consistent labeling of
the data. In fact, once stability is reached, all players play the strategy with
the highest payoff. Each player arrives to this state not only considering its
own strategies but also the strategies that its co-players are playing. For each
player $i \in I$ is chosen the strategy with the highest probability when
the system converges (see equation below).%
\begin{equation}\label{eq:max} \phi_i = \argmax{h=1,\dots,c}x_{ih}
\end{equation} \noindent In our framework a word is not disambiguated only if it
is not able to update its strategy space. This can happen when the player's
strategy space is initialized with a uniform distribution and either its payoff
matrices have only zero entries, that is when its senses are not similar to the
senses of the co-players, or it is not connected with other nodes in the graph.
The former assumption depends on the semantic measures used to calculate the
payoffs (see section \ref{sec:semMes}), experimentally we noticed that it does
not happen frequently. The latter assumption can happen when a word is not
present in a determined corpus. It can be avoided using query expansion
techniques or connecting the disconnected node with nodes in its neighborhood,
exploiting proximity relations (see section \ref{sec:grCns}). With equation
\ref{eq:max} it is guaranteed that at the end of the process each word is mapped
to exactly one sense. Experimentally, we noticed that when a word is able to
update its strategy space, it is not the case that two strategies in it have the
same probability.

\subsection{Implementation of the WSD Games} In order to run our algorithm we need
the network that models the interactions among the players, the strategy space
of the game and the payoff matrices. We adopted the following steps in order to
model the data required by our framework and specifically, for each text to be
disambiguated, we: \begin{itemize}
  \item extract from the text the list of words $I$, which have an entry in a
        lexical database,
  \item compute, from $I$, the word similarity matrix $W$ in which are stored
        the pairwise similarities among each word with the others and represents
        the players' interactions,
\item increase the weights between two words, which share a proximity relation,
  \item extract, from $I$, the list $C$ of all the possible senses, which
        represents the strategy space of the system,
  \item assign, for each word in $I$, a probability distribution over the senses
        in $C$ creating for each player a probability distribution over the
        possible strategies,
  \item compute the sense similarity matrix $Z$ among each pair of senses in
        $C$, which is then used to compute the partial payoff matrices of each 
        games,
   \item apply the replicator dynamics equation in order to compute the Nash 
   equilibria of the games, and
   \item assign to each word $i \in I$ a strategy $s \in C$. \end{itemize}%
\noindent These steps are described in the following section. In Section
\ref{sec:grCns} we describe the graph construction procedure, which we employed
in order to model the geometry of the data. In Section \ref{sec:sspace} we
explain how we implement the strategy space of the game, which allows each
player to choose over a predetermined number of strategies. In Section
\ref{sec:wsim} we describe how we compute the sense similarity matrix and how it
is used to create the partial payoff matrices of the games. Finally in Section
\ref{sec:sysDyn} we describe the system dynamics.%
\subsubsection{Graph Construction}\label{sec:grCns} In our study, we modeled the
geometry of the data as a graph. The nodes of the graph correspond to the words
of a text, which have an entry in a lexical database. We denote the words by
$I=\{i_j\}_{j=1}^N$, where $i_j$ is the $j$-th word and $N$ is the total number
of words retrieved. From $I$ we construct a $N \times N$ similarity matrix $W$
where each element $w_{ij}$ is the similarity value assigned by a similarity
function to the words $i$ and $j$. $W$ can be exploited as an useful tool for
graph-based algorithms since it is treatable as weighted adjacency matrix of a
weighted graph. 

 A crucial factor for the graph construction is the choice
of the similarity measure, $sim(\cdot,\cdot) \rightarrow \mathbb{R}$ to weight
the edges of the graph. In our experiments, we used similarity measures, which
compute the strength of co-occurrence between any two words $i_i$ and $i_j$.
\begin{equation} w_{ij} = sim(i_i,i_j) \text{ } \forall i,j \in I:i \neq j
\end{equation} \noindent This choice is motivated by the fact that collocated
words tend to have determined meanings \cite{gale1992method,yarowsky1993one},
and also because the computation of these similarities can be obtained easily.
In fact, it only required a corpus in order to compute a vast range of
similarity measures. Furthermore, large corpora such as the BNC corpus
\cite{leech1992100} and the Google Web 1T corpus \cite{brants2006web} are freely
available and extensively used by the research community. 

 In some cases, it
is possible that some target words are not present in the reference corpus, due
to different text segmentation techniques or spelling differences. In this case
we use query expansion techniques in order to find an appropriate substitute
\cite{carpineto2012survey}. Specifically, we use WordNet to find alternative
lexicalizations of a lemma, choosing the one that co-occurs more frequently
with the words in its context. 

 The information obtained from an association
measure can be enriched taking into account the proximity of the words in the
text (or the syntactic structure of the sentence). The first task can be
achieved augmenting the similarities among a target word and the $n$ words that
appear on its right and on its left, where $n$ is a parameter that with small
values can capture fixed expressions and with large values can detect semantic
concepts \cite{fkih2012learning}. The second task can be achieved using a
dependency parser to obtain the syntactical relations among the words in the
target sentence, but this approach is not used in this paper. In this way, the
system is able to exploit local and global cues, mixing together the \emph{one
sense per discourse} \cite{kelly1975computer} and the \emph {one sense per
collocation} \cite{yarowsky1993one} hypotheses.

%

We are not interested in all the relations in the sentence but we focus only on
relations among target words. The use of a dependency/proximity structure makes
the graph reflect the structure of the sentence while the use of a
distributional approach allows us to exploit the relations of semantically
correlated words. This is particularly useful when the proximity information is
poor; for example when it connects words to auxiliary or modal verbs.
Furthermore, these operations ensure that there are no disconnected nodes in the
graph.%
\subsubsection{Strategy Space Implementation}\label{sec:sspace} The strategy
space of the game is created using a knowledge base to collect the sense
inventories $M_i=\{1,\dots,m_i\}$ of each word in a text, where $m_i$ is the
number of senses associated to word $i$. Then is created the list
$C=(1,\dots,c)$ of all the unique concepts in the sense inventories, which
correspond to the space of the game. 

 With this information we can define
the strategy space $S$ of the game in matrix form as:%
\begin{center}
$ \begin{matrix} s_{i1} & s_{i2} & \cdots
& s_{ic} \\ \vdots & \vdots & \cdots & \vdots \\ s_{n1} & s_{n2} & \cdots &
s_{nc} \end{matrix} $
\end{center}%
\noindent where each row correspond to the mixed strategy space of a player and
each column correspond to a specific sense. Each component $s_{ih}$ denotes the
probability that the player chooses to play its $h$th pure strategy among all
the strategies in its strategy profile, as described in Section \ref{sec:GT}.
The initialization of each mixed strategy space can either be uniform or take into
account information from sense-labeled corpora.

\subsubsection{The Payoff Matrices}\label{sec:wsim}

We encoded the payoff matrix of a WSD game as a sense similarity matrix among
all the senses in the strategy spaces of the game. In this, way the higher the
similarity among the senses of two words, the higher the incentive for a
word to chose that sense, and play the strategy associated with it. 

 The $c
\times c$ sense similarity matrix $Z$ is defined in equation (\ref{eq:wsim}).%
\begin{equation}\label{eq:wsim}
  z_{ij}=ssim(s_i,s_j) \text{ } \forall i,j \in C:i \neq j
\end{equation}%
\noindent This similarity matrix can be obtained using the information derived
by the same knowledge base used to construct the strategy space of the game. It
is used to extract the partial payoff matrix $Z_{ij}$ for all the single games
played between two players $i$ and $j$. This operation is done extracting from $Z$
the entries relative to the indices of the senses in the sense inventories $M_i$
and $M_j$. It produces an $m_i \times m_j$ payoff matrix, where $m_i$ and $m_j$
are the numbers of senses in $M_i$ and $M_j$, respectively.%
\subsubsection{System Dynamics}\label{sec:sysDyn} Now that we have the topology
of the data $W$, the strategy space of the game $S$ and the payoff matrix $Z$ we
can compute the Nash equilibria of the game according to equation
(\ref{eq:replicatorDS}). In each iteration of the system each player plays a
game with its neighbors $N_i$ according to the co-occurrence graph $W$. The payoffs
of the \textit{h-th} strategy is calculated as:
\begin{equation}\label{eq:payoff} u_i(e^h,x)=\sum_{j \in N_i}(w_{ij}Z_{ij}x_j)_h
\end{equation}%
\noindent and the player's payoff as:%
\begin{equation}
u_i(x) =\sum_{j \in N_i} x_i^T (w_{ij}Z_{i j}x_j)
\end{equation}%
\noindent%
In this way we can weight the influence that each word has on the choices that a
particular word has to make on its meaning. We assume that the payoff of word
$i$ depends on the similarity that it has with word $j$, $w_{ij}$, the
similarities among its senses and those of word $j$, $Z_{ij}$, and the sense
preference of word $j$, ($x_j$). During each phase of the dynamics a process of
selection allows strategies with higher payoff to emerge and at the end of the
process each player chooses its sense according to these constraints. 

 The
complexity of each step of the replicator dynamics is quadratic but there are
different dynamics that can be used with our framework to solve the problem more
efficiently, such as the recently introduced \emph{infection and immunization}
dynamics \cite{bulo2011graph} that has a linear-time/space complexity per step and it is 
known to be much faster then, and as accurate as, the replicator dynamics.
\subsection{Implementation Details}\label{sec:impDetails} In this section we describe the
association measures used to weight the graph $W$ (Section
\ref{sec:disMes}), the semantic and relatedness measures used to compare the
synsets (Section \ref{sec:semMes}), the computation of the payoff matrices of
the games (Section \ref{sec:simPay}) and the different implementations of the
system strategy space (\ref{sec:strSpImp}), in case of unsupervised,
semi-supervised and coarse-grained WSD.%
\subsubsection{Association Measures}\label{sec:disMes} We
evaluated our algorithm with different similarity measures in order to find the
measure that performs better. The results of this evaluation are presented in
Section \ref{sec:evalRisUns}. Specifically for our experiments we used eight
different measures: the \emph{Dice coefficient} (\emph{dice})
\cite{dice1945measures}, the \emph{modified Dice coefficient} (\emph{mDice})
\cite{kitamura1996automatic}, 
the pointwise mutual information (\emph{pmi}) \cite{church1990word},
the \emph{t-score} measure (\emph{t-score}) \cite{church1990word}, the
\emph{z-score} measure (\emph{z-score}) \cite{burrows2002delta}, the \emph{odds
ration} (\emph{odds-r}) \cite{blaheta2001unsupervised}, the \emph{chi-squared}
test (\emph{chi-s}) \cite{rao2002karl} and the \emph{chi-squared} correct
(\emph{chi-s-c}) \cite{degroot1986probability}. 

 The measures that we used
are presented in Figure \ref{fig:meas} where the notation refers to the standard
contingency tables \cite{evert2008corpora} used to display the observed and
expected frequency distribution of the variables, respectively on the left and
on the right of Figure \ref{fig:contingency}. All the measures for the
experiments in this article have been calculated using the BNC corpus
\cite{leech1992100}, since it is a well balanced general domain corpus.%
\begin{figure}
  \[\arraycolsep=7.0pt\def\arraystretch{1.3}
  \begin{array}{lll}
   \begin{array}{l|ll|l}
   & w_j & \neg w_j &  \\ 
  w_i & O_{11} & O_{12} & = R_1 \\ 
  \neg w_i & O_{21} & O_{22} & = R_2 \\ \hline
  & = C_1 & = C_2 & = N 
    \end{array} 
    &
    \text{ } \text{ } \text{ }\text{ }
    &
       \begin{array}{c|c|c}
   & w_j & \neg w_j   \\ \hline
  w_i & E_{11} = R_1C_1/N &  E_{12} = R_1C_2/N  \\  \hline
  \neg w_i &  E_{21} = R_2C_1/N &  E_{22} = R_2C_2/N  \\ \hline
 \multicolumn{1}{l}{ } & \multicolumn{1}{l}{ } & \\
    \end{array} 
      \end{array} 
      \]
\caption{Contingency tables of observer frequency (on the left)
and expected frequency (on the right).}\label{fig:contingency}
\end{figure}%
\begin{figure}
\[\arraycolsep=12.0pt\def\arraystretch{2.5}
   \begin{array}{lll}
    \text{\emph{dice}} = \frac{2O_{11}}{R_1+C_1} & \text{\emph{m-dice}} =
log_2O_{11}\frac{2O_{11}}{R_1+C_1} & pmi = log_2 \frac{0}{E_{11}}  \\
  \text{\emph{t-score}} = \frac{O-E_{11}}{\sqrt{O}} &
  \text{\emph{odds-r}}= log \frac{(O_{11}+1/2)(O_{22}+1/2)}{(O_{12}+1/2)(O_{21}+1/2)} & \text{\emph{z-score}}=\frac{O-E_{11}}{\sqrt{E_{11}}} \\
    \text{\emph{chi-s}} = \sum_{ij}\frac{(O_{ij}-E_{ij})^2}{E_{ij}} & \text{\emph{chi-s-c}} = \frac{N(|O_{11}O_{22}-O_{12}O_{21}|-N/2)^2}{R_1R_2C_1C_2} & 
    \\ \hline
  \end{array}
  \]
  \caption{Association measures used to weight the co-occurrence graph $W$.}\label{fig:meas}
  \end{figure}
\subsubsection{Semantic and Relatedness Measures}\label{sec:semMes}
We used WordNet \cite{miller1995wordnet} and BabelNet \cite{navigli2012babelnet}
as knowledge bases to collect the sense inventories of each word to be
disambiguated.%

agraph{Semantic and Relatedness Measures Calculated with
WordNet}\label{sec:semMesWN} WordNet \cite{miller1995wordnet} is a lexical
database where the lexicon is organized according to a psycholinguistic theory
of the human lexical memory, in which the vocabulary is organized conceptually
rather than alphabetically, giving a prominence to word meanings rather than to
lexical forms. The database is divided in five parts: nouns, verbs, adjectives,
adverbs and functional words. In each part the lexical forms are mapped to the
senses related to them, in this way it is possible to cluster words, which share
a particular meaning (synonyms) and to create the basic component of the
resource: the \emph{synset}. Each \emph{synset} is connected in a network to
other synsets, which have a semantic relation with it.

 The relations in
WordNet are: hyponymy, hypernymy, antonymy, meronymy and holonymy. Hyponymy
gives the relations from more general concepts to more specific; hypernymy gives
the relations from particular concepts to more general; antonymy relates two
concepts, which have an opposite meaning; meronymy connects the concept that is
part of a given concept with it; and holonymy relates a concept with its
constituents. Furthermore, each synset is associated to a definition and gives
the morphological relations of the word forms related to it. Given the
popularity of the resource many parallel projects have been developed. One of
them is \emph{eXtended WordNet} \cite{mihalcea2001extended}, which gives a
parsed version of the glosses together with their logical form and the
disambiguation of the term in it. 

 We have used this resource to compute
similarity and relatedness measures in order to construct the payoff matrices of
the games. The computation of the sense similarity measures is generally
conducted using relations of likeness such as the \emph{is-a} relation in a
taxonomy; on the other hand the relatedness measures are more general and take
in account a wider range of relations such as the \emph{is-a-part-of} or
\emph{is-the-opposite-of}.

 The semantic similarity measure which we used are
the \emph{wup similarity} \cite{wu1994verbs} and the \emph{jcn measure}
\cite{jiang1997semantic}. These measure are based on the structural organization
of WordNet and compute the similarity among two senses $s_i$, $s_j$ according to
the depth of the two sense in the lexical database and that of the most specific
ancestor node, \emph{msa}, of the two senses. The \emph{wup similarity},
described in equation (\ref{eq:wup}), takes into account only the path length
among two concepts. The \emph{jcn measure} combines corpus statistics and
structural properties of a knowledge base. It is computed as presented in
equation (\ref{eq:jcn}), where $IC$ is the information content of a concept $c$
derived from a corpus\footnote{We used the IC files computed on SemCor
\cite{miller1993semantic} for the experiments in this article. They are
available at http://wn-similarity.sourceforge.net and are mapped to the
corresponding version of WordNet of each dataset.} and computed as
$IC(c)=log^{-1}P(c)$. \begin{equation}\label{eq:wup} ssim_{wup}(s_i,s_j) =
2*depth(msa) / (depth(s_i) + depth(s_j)) \end{equation} \noindent
\begin{equation}\label{eq:jcn} ssim_{jcn}(s_i,s_j) =IC(s_1) + IC(s_2) - 2
IC(msa) \end{equation}

The semantic relatedness measures, which we used, are based on the computation
of the similarity among the definitions of two concepts in a lexical database.
These definitions are derived from the glosses of the synsets in WordNet. They
are used to construct a co-occurrence vector $v_i=(w_{1,i},w_{2,i} ... w_{n,i})$
for each concept $i$, with a bag-of-words approach where $w$ represents the
number of times word $w$ occur in the gloss and $n$ is the total number of
different words (\emph{types}) in the corpus\footnote{In our case the corpus is
composed of all the WordNet glosses.}. This representation allows to project
each vector into a \emph{vector space} where it is possible to conduct different
kind of computations. For our experiments, we decided to calculate the
similarity among two glosses using the cosine distance among two vectors as
shown in equation (\ref{eq:cosine}), where the nominator is the intersection of
the words in the two glosses and $||v||$ is the norm of the vectors, which is
calculated as: $ \sqrt{\sum_{i=1}^{n}w_i^2} $.%
\begin{equation}\label{eq:cosine}
\cos \theta \frac{v_i \cdot v_j}{||v_i|| ||v_j||}
\end{equation}%
\noindent This measure gives the cosine of the angle
between the two vectors and in our case returns values ranging from $0$ to $1$
because the values in the co-occurrence vectors are all positive. Given the fact
that small cosine distances indicate a high similarity we transform this
distance measure into a similarity measure with $1-cos(v_i,v_j)$. 

 The
procedure to compute the semantic relatedness of two synsets has been introduced
by Patwardhan and Pedersen \shortcite{patwardhan2006using} as \emph{Gloss Vector
measure} and we used it with four different variations for our experiments. The
four variations are named: $tf-idf$, $tf\-idf_{ext}$, $vec$ and
$vec_{ext}$. The difference among them relies on the way the gloss
vectors are constructed. Since the synset gloss is usually short we used the
concept of \emph{super-gloss} as in \cite{patwardhan2006using} to construct the
vector of each synset. A \emph{super-gloss} is the concatenation of the gloss of
the synset plus the glosses of the synsets, which are connected to it via some
WordNet relations \cite{pedersen2012duluth}. We employed, the WordNet version
that has been used to to label each dataset. Specifically the different
implementations of the vector construction vary on: the way in which the
co-occurrence is calculated, the corpus used and the source of the relations.
\emph{tf-idf} constructs the co-occurrence vectors exploiting the \emph{term
frequency - inverse document frequency} weighting schema (\emph{tf-idf}).
$tf-idf_{ext}$ uses the same information of \emph{tf-idf} plus
the relations derived from eXtended WordNet \cite{mihalcea2001extended}. %
\emph{vec} uses a standard BoW approach to compute the co-occurrences. $vec_{ext}$ uses the same information of \emph{vec} plus the relations from eXtended WordNet.%

 Instead of considering only the raw frequency of terms in documents, the
\emph{tf-idf} method, scales the importance of less informative terms taking
into account the number of documents in which a term occur. Formally, it is the
product of two statistics: the term frequency and the inverse document
frequency. The former is computed as the number of times a term occur in a
document (gloss in our case), the latter is computed as $idf_t = \log
\frac{N}{df_t}$, where $N$ is the number of documents in the corpus and $df_t$
is the number of documents in which the term occurs.%

agraph{Relatedness Measure Calculated with BabelNet and NASARI}\label{sec:semMesBN}%
BabelNet \cite{navigli2012babelnet} is a wide-coverage multilingual semantic
network. It integrates lexicographic and encyclopedic knowledge from WordNet and
Wikipedia, automatically mapping the concepts shared by the two knowledge bases.
This mapping generates a semantic network where millions of concepts are
lexicalized in different languages. Furthermore, it allows to link \emph{named
entities}, such as \emph{Johann Sebastian Bach} and concepts, such as
\emph{composer} and \emph{organist}.

 BabelNet can be represented as a
labeled direct graph $G = (V , E)$ where $V$ is the set of nodes
(\emph{concepts} or \emph{named entities}) and $E \subseteq V \times R \times V$
is the set of edges connecting pairs of \emph{concepts} or \emph{named
entities}. The edges are labeled with a semantic relation from $R$, such as:
\emph{is-a}, \emph{given name} or \emph{occupation}. Each node $v \in V$
contains a set of lexicalizations of the concept for different languages, which
forms a BabelNet synset.

 The semantic measure, which we developed using
BabelNet, is based on NASARI\footnote{The resource is available at
http://lcl.uniroma1.it/nasari/} \cite{camacho2015nasari}, a semantic
representation of the \emph{concepts} and \emph{named entities} in BabelNet.
This approach first exploits the BabelNet network to find the set of related
\emph{concepts} in WordNet and Wikipedia and then constructs two vectors to
obtain a semantic representation of a concept $b$. These representations are
projected in two different semantic spaces, one based on words and the other on
synsets. They use lexical specificity\footnote{A statistical measure based on
the hypergeometric distribution over word frequencies.}
\cite{lafon1980variabilite} to extract the most representative words to use in
the first vector and the most representative synsets to use in the second
vector. 

 In this article, we computed the similarity between two senses using
the vectors (of the word-based semantic space) provided by NASARI. These
semantic representations provide for each sense the set of words, which best
represent e particular concept and the score of representativeness of each word.
From this representation we computed the pairwise cosine similarity between each
concept as described in the previous section for the semantic relatedness
measures.

 The use of NASARI is particularly useful in case of named entity
disambiguation, since it includes many entities, which are not included in
WordNet. On the other hand, it is difficult to use it in all-words sense
disambiguation tasks, since it includes only WordNet synsets that are mapped to
Wikipedia pages in BabelNet. For this reason it is not possible to find the
semantic representation for many verbs, adjectives and adverbs, which is common
to find in all-words sense disambiguation tasks.

 We used the SPARQL
endpoint\footnote{http://babelnet.org/sparql/} provided by BabelNet to collect
the sense inventories of each word in the texts of each dataset. For this task
we filtered the first 100 resources whose label contains the lexicalization of
to word to be disambiguated. This operation is required because in many cases it
is possible to have indirect references to entities.
%
%
%
\subsubsection{From similarities to payoffs}\label{sec:simPay} The similarity and relatedness
measures are computed for all the senses of the words to be disambiguated. From
this computation it is possible to obtain a similarity matrix $Z$, which
incorporates the pairwise similarity among all the possible senses. This
computation could have heavy computational cost, if there are many words to be
disambiguated. To overcome this issue, the pairwise similarities can be computed
just one time on the entire knowledge base and used in actual situations,
reducing the computational cost of the algorithm. From this matrix we can obtain the
partial semantic similarity matrix for each pair of player, $Z_{ij} = m \times n
$, where $m$ and $n$ are the senses of $i$ and $j$ in $Z$.%

 In a previous work \cite{tripodi2015semeval} we did not use this information, instead we used labeled data points to propagate the class membership information over the graph. In this new version the use of the semantic information made the algorithm completely unsupervised.
\subsubsection{Strategy space implementation}\label{sec:strSpImp} Once the
pairwise similarities between the words and their senses, stored in the two
matrices $W$ and $Z$, are calculated, we can pass to the description of the
strategy space of each player. It can be initialized with equation
(\ref{eq:normaldist}), which follows the constraints described in Section
\ref{sec:EGT} and assigns to each sense an equal probability.
\begin{equation}\label{eq:normaldist}
s_{ij}=\begin{cases} |M_i|^{-1}, & \text{if sense $ j $ is in $ M_i$}.\\ 0, &
\text{otherwise}. \end{cases} \end{equation} \noindent This initialization is
used in the case of unsupervised WSD since it does not use any prior knowledge
about the senses distribution. In case we want to exploit information from prior
knowledge, obtained from sense-labeled data, we can assign to each sense a
probability according with its rank, concentrating a higher probability on
senses that have a high frequency. To model this kind of scenario we used a
geometric distribution that gives us a decreasing probability distribution.
This new initialization is defined as follows,
\begin{equation}\label{eq:geodist} s_{ij}=\begin{cases} p(1-p)^{r_j}, & \text{if
sense $ j $ is in $ M_i$}.\\ 0, & \text{otherwise}. \end{cases} \end{equation}
\noindent where $p$ is the parameter of the geometric distribution and
determines the scale or statistical dispersion of the probability distribution,
and $r_j$ is the rank of sense $j$, which ranges from $1$, the rank of the most
common sense, to $m$, the rank of the least frequent sense. Finally, the vector
obtained from equation (\ref{eq:geodist}) is divided by $\sum_{j \in S_i}p_j$ in
order to make the probabilities add up to $1$. In our experiments, we used the
ranked system provided by the Natural Language Toolkit (version 3.0)
\cite{bird2006nltk} to rank the senses associated to each word to be
disambiguated. Natural Language Toolkit is a suite of modules and data sets,
covering symbolic and statistical NLP. It includes a WordNet reader that can be
queried with a lemma and a part of speech to obtain the list of possible sysnets
associated to the specified lemma and a part of speech. The returned synsets are
listed in decreasing order of frequency and can be used as ranking system by our
algorithm. 

 We used the method proposed by Navigli
\shortcite{navigli2006meaningful} for the experiments on coarse-grained WSD.
With this approach it is possible to cluster the senses of a given word,
according to the similarity that the senses share. In this way it is possible to
obtain a set of disjoint clusters $O = \{o_1,..., o_t \}$, which is ranked
according to the information obtained with the ranking system described above,
for each sense inventory $M$. The initialization of the strategy space, in this
case, is defined as follows, \begin{equation}\label{eq:geodistClust}
s_{ij}=\begin{cases} p(1-p)^{r_o}, & \text{if sense $ j $ is in cluster $o$}.\\
0, & \text{otherwise}. \end{cases} \end{equation} \noindent With this
initialization it is possible to assign an equal probability to the senses
belonging to a determined cluster and to rank the clusters according to the
ranking of the senses in each of them.%
\subsection{An example}\label{sec:esempio}
\begin{figure}[ht!] \begin{center}
        \subfigure[Co-occurrence graph]{%
            \label{fig:first}
            \includegraphics[width=0.31\textwidth]{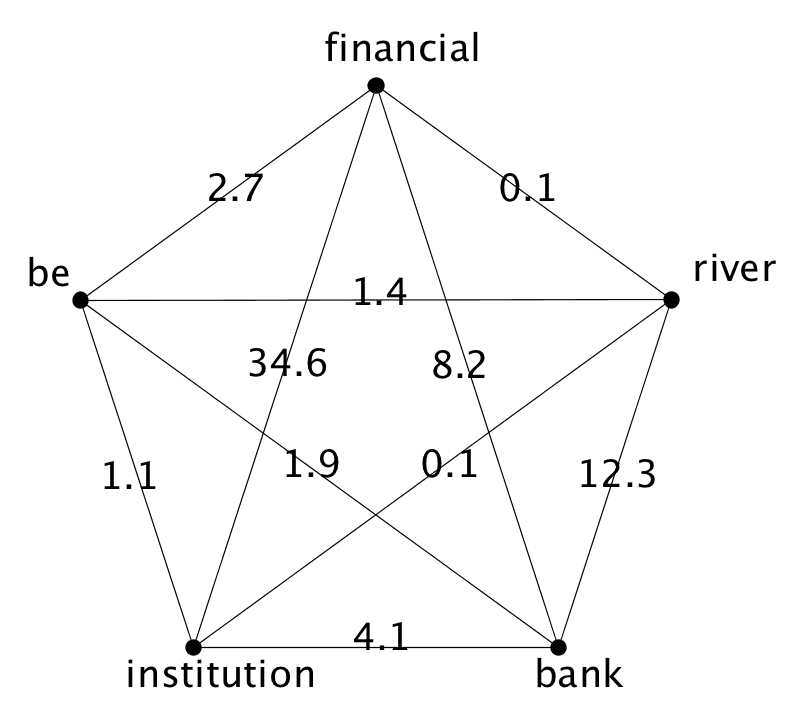} 
        }%
        \subfigure[n-gram graph]{%
           \label{fig:third}
           \includegraphics[width=0.31\textwidth]{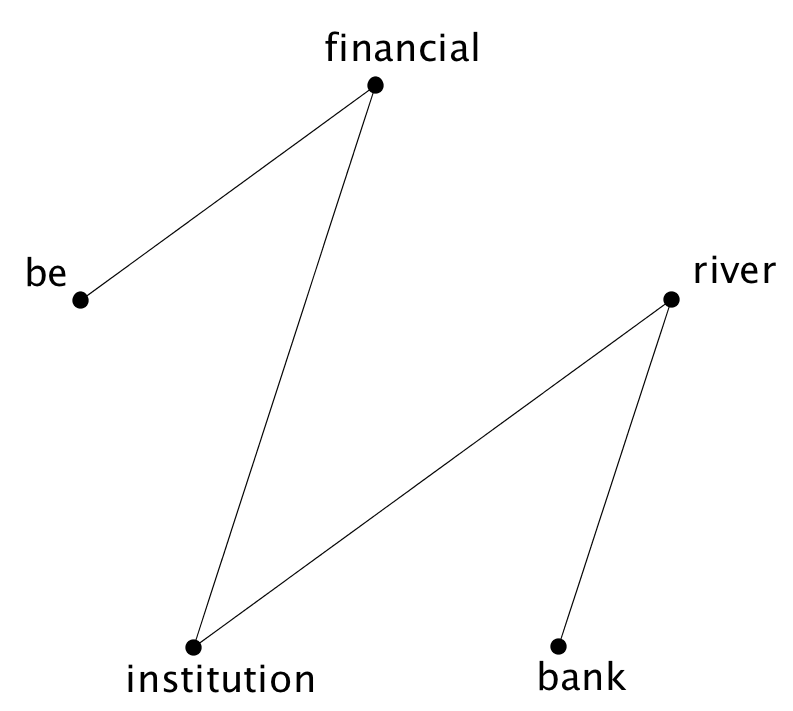} 
        }
        \subfigure[Similarity n-gram graph]{%
            \label{fig:fourth}
            \includegraphics[width=0.31\textwidth]{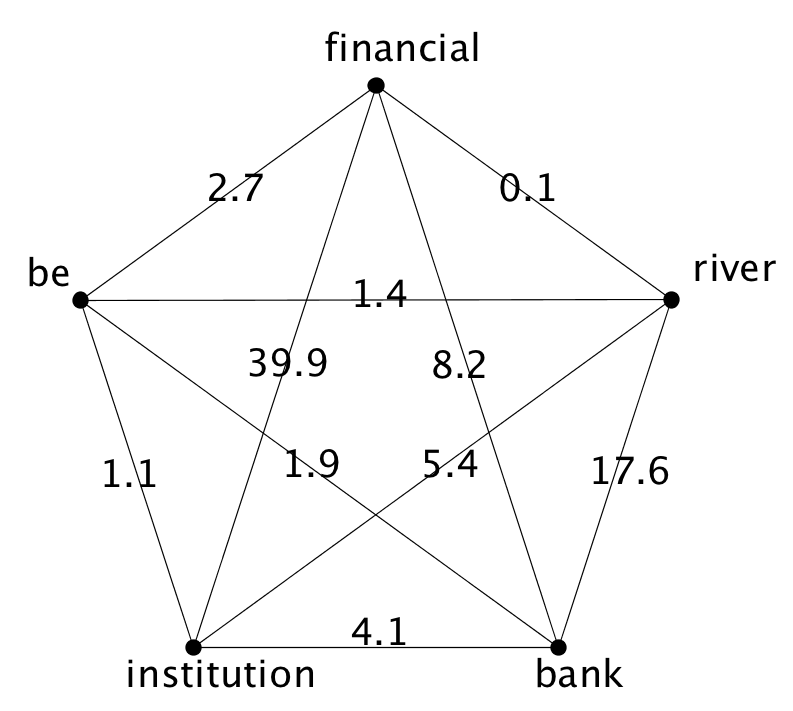} 
        }%
    \end{center}
    \caption{%
 Three graph representations for the sentence: \textit{there is a financial
 institution near the river bank}. \protect\subref{fig:first} a co-occurrence
 graph constructed using the modified Dice coefficient as similarity measure
 over the the Google Web 1T 5-Gram Database \protect\cite{brants2006web} to
 weight the edges. 
 \protect\subref{fig:third} graph representation of the n-gram structure of the sentence, with $n=1$; for each
 node, an edge is added to another node if the corresponding word appears to its
 left or right, in a window of size one word. \protect\subref{fig:fourth} a
 weighted graph that combines the information of the co-occurrence graph and the
 n-gram graph. The edges of the co-occurrence graph are augmented by its mean
 weight if a corresponding edge exists in the n-gram graph and not include a
 stop-word.}%
\label{fig:subfigures} \end{figure}

As an example we can consider the following sentence, which we encountered before:
\begin{itemize}
  \item{There is a financial institution near the river bank.} \end{itemize}%
\noindent We first tokenize, lemmatize and tag the sentence, then we extract the
content words that have an entry in WordNet 3.0 \cite{miller1995wordnet},
constructing the list of words to be disambiguated: \{is, financial,
institution, river, bank\}. Once we identified the target words we computed the
pairwise similarity for each target word. For this task we used the Google Web
1T 5-Gram Database \cite{brants2006web} to compute the modified Dice
coefficient\footnote{Specifically we used the service provided by the Corpus
Linguistics group at FAU Erlangen-N\"{u}rnberg, with a collocation span of 4
words on the left and on the right and collocates with minimum frequency:
100.} \cite{kitamura1996automatic}. With the information derived by this process
we can construct a co-occurrence graph (Figure \ref{fig:first}), which indicates
the strength of association between the words in the text. This information can
be augmented taking into account other sources of information such that the
dependency structure of the syntactic relations between the words\footnote{This
aspect is not treated in this article.} or the proximity information derived by
a simple n-gram model (Figure \ref{fig:third}, $n=1$). 

 The operation to increment
the weights of structurally related words is important because it prevents the
system to rely only on distributional information, which could lead to a sense
shift for the ambiguous word \emph{bank}. In fact, its association with the
words \emph{financial} and \emph{institution} would have the effect to interpret
it as a \emph{financial institution} and not as \emph{sloping land} as defined
in WordNet. Furthermore, using only distributional information could exclude
associations between words that do not appear in the corpus in use.

 In
Figure \ref{fig:fourth} it is represented the final form of the graph for our
target sentence, in which we have combined the information from the
co-occurrence graph and from the n-gram graph. The weights in the co-occurrence
graph are increased by the mean weight of the graph if a corresponding edge
exists in the n-gram graph and not include stop-word\footnote{A more accurate
representation of the data can be obtained using the dependency structure of the
sentence, instead of the n-gram graph; but in this case the results would not
have changed, since in both cases there is an edge between \emph{river} and
\emph{bank}. In fact, in many cases a simple n-gram model can implicitly detect
syntactical relations. We used the stop-word list available in the Python
Natural Language Toolkit, described above.}.
  
\begin{figure} \includegraphics[width=1\textwidth]{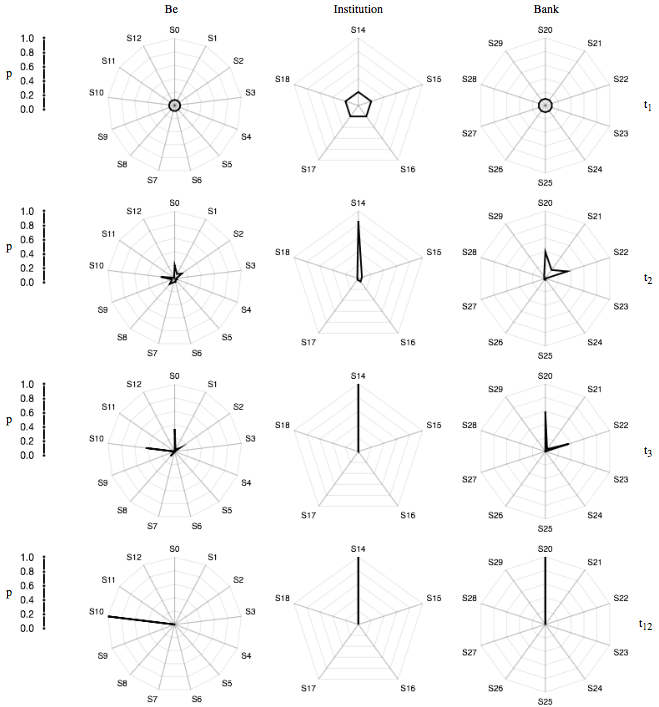} 
\caption{System dynamics for the words: \emph{be},
\emph{institution} and \emph{bank} at time step 1,2,3 and 12 (system
convergence). The strategy space of each word is represented as a regular
polygon of radius 1, where the distance from the center to any vertex represents
the probability associated to a particular word sense. The values on each radius
in a polygon are connected with a darker line in order to show the actual
probability distribution obtained at each time step. }
\label{fig:dynamics} \end{figure}%
After the pairwise similarities between the words are computed we access a
lexical database in order to get the sense inventories of each word so that each
word can be associated to a predefined number of senses. For this task, we use
WordNet 3.0 \cite{miller1995wordnet}. Then for each unique sense in all the
sense inventories we compute the pairwise semantic similarity, in order to
identify the affinity among all the pairwise sense combination. This task can be
done using a semantic similarity or relatedness measure \footnote{Semantic
similarity and relatedness measures are discussed in Section \ref{sec:disMes}
and \ref{sec:semMes}.}. For this example, we used a variant of the \emph{gloss
vector measure} \cite{patwardhan2006using}, the \emph{tf-idf}, described in
Section \ref{sec:semMes}. 

 Having obtained the similarity information we can
initialize the strategy space of each player with a uniform distribution, given
the fact that we are not considering any prior information about the senses
distributions. Now the system dynamics can be started. In each iteration of the
dynamics each player play a game with its neighbors obtaining a payoff for each
of its strategies according to equation (\ref{eq:payoff}) and once the players
have played the games with their neighbors in $W$, the strategy space of each
player is updated at time $t+1$ according to equation (\ref{eq:replicatorDS}).

 We present the dynamics of the system created for the example sentence in
Figure \ref{fig:dynamics}. The dynamics are shown only for the ambiguous words
at time steps $t_1$, $t_2$, $t_3$ and $t_{12}$ (when the system converges). As
we can see at time step 1 the senses of each word are equiprobable, but as soon
as the games are played some sense starts to emerge. In fact at time step 2 many
senses are discarded, and this in virtue of two principles,%
\begin{enumerate*}[label=\itshape\alph*\upshape)]%
\item the words in the text push the senses of the other words      
      toward a specific sense; and%
\item the sense similarity values for certain senses are very low.
      \end{enumerate*}%
\noindent
Regarding the first principle, we can consider the word \emph{institution},
which is playing the games with the words \emph{financial} and \emph{bank}, is
immediately driven toward a specific sense, as an organization founded and
united for a specific purpose as defined in WordNet 3.0; thus discarding the
other senses. Regarding the second principle, we can consider many senses of the
word \emph{bank} that are not compatible with the senses of the other words in
the text and therefore their values decrease rapidly.

 The most interesting
phenomenon that can be appreciated from the example is the behavior of the
strategy space of the word \emph{bank}. It has ten senses, according to WordNet
3.0 \cite{miller1995wordnet}, and can be used in different context and domains,
to indicate, among the other things, a financial institution ($s_{22}$ in Figure
\ref{fig:dynamics}) or a sloping land ($s_{20}$ in Figure \ref{fig:dynamics}).
When it plays a game with the words \emph{financial} and \emph{institution} it
is directed toward its financial sense; when it plays a game with the word
\emph{river}, it is directed toward its naturalistic meaning. As we can see in
Figure \ref{fig:dynamics} at time step 2 the two meanings ($s_{20}$ and
$s_{22}$) have almost the same value and at time step 3 the word starts to
define a precise meaning to the detriment of $s_{21}$ but not of $s_{22}$. The
balancing of these forces toward a specific meaning is given by the similarity
value $w_{ij}$, which allows \emph{bank} in this case to chose its naturalistic
meaning. Furthermore, we can see that the inclination to a particular sense is
given by the payoff matrix $Z_{ij}$ and by the strategy distribution $S_j$,
which indicates what sense word $j$ is going to choose, ensuring that word $i$'s
is coherent with this choice.%
%
%
%
\section{Experimental Evaluation}\label{sec:eval}%
 In this Section we describe how the parameters of the presented method have
 been found and how it has been tested and compared with state-of-the-art
 systems\footnote{The code of the algorithm and the datasets used are available
 at http://www.dsi.unive.it/~tripodi/wsd}, in Section \ref{sec:tuning} and
 Section \ref{sec:evalSetup}, respectively. We describe the datasets used for
 the tuning and for the evaluation of our model and the different settings used
 to test it. The results of our experiments using WordNet as knowledge base are
 described in Section \ref{sec:evalRisUns}, where two different implementations
 of the system are proposed, the unsupervised and the supervised. In Section
 \ref{sec:evalSota} we compare our results with state-of-the-art systems.
 Finally, the results of the experiments using BabelNet as knowledge base,
 related to WSD and entity disambiguation, are described in Section \ref{sec:evalBabel}. The results are provided as F1, computed according to the following equation,%
 \begin{equation}\label{eq:f1}%
   F1 = 2 \cdot \frac{precision \cdot recall}{precision + recall} \cdot 100.%
   \end{equation}%
\noindent F1 is a measure that determines the weighted harmonic mean of precision and recall. Precision is defined as the number of correct answers divided by the number of provided answers and recall is defined as the number of correct answers divided by the total number of answers to be provided.%
\subsection{Parameter Tuning}\label{sec:tuning} We used two datasets to tune the
parameters of our approach, SemEval-2010 task 17 (S10) \cite{agirre2009semeval}
and SemEval-2015 Task 13 (S15) \cite{moro2015semeval}. The first dataset is
composed of three English texts from the ecology domain, for a total of 1398
words to be disambiguated (1032 nouns/named entities and 366 verbs). The second
dataset is composed of four English documents, from different domains: medical,
drug, math and social issues, for a total of 1261 instances, including
nouns/named entities, verbs, adjectives and adverbs. Both datasets have been
manually labeled using WordNet 3.0. The only difference between these datasets is
that the target words of the first dataset belong to a specific domain, whereas
all the content words of the second dataset have to be disambiguated. We used
these two typologies of dataset to evaluate our algorithm in different scenarios,
furthermore we created, from each dataset, 50 different datasets, selecting from
each text a random number of sentences and evaluating our approach on each of
these datasets to identify the parameters that on average perform better than
others. In this way it is possible to simulate a situation in which the system
has to work on texts of different sizes and on different domains. This because
as demonstrated by S{\o}gaard et. al \shortcite{sogaard2014s} the results of a
determined algorithm are very sensitive to sample size. The number of target
words for each text in the random datasets ranges from 12 to 571. The parameters
which will be tuned are: the association and semantic measure to use to weight
the similarity among words and senses (Section \ref{sec:parTunMeas}), the $n$ of
the n-gram graph used to increase the weights of near words (Section
\ref{sec:parTunK}) and the $p$ of the geometric distribution useed by our
semi-supervised system (Section \ref{sec:parTunP})%
\subsubsection{Association
and Semantic Measures}\label{sec:parTunMeas} The first experiment that we
present is aimed at finding the semantic and distributional measures with the
highest performances. We recall that we used WordNet 3.0 as knowledge base and
the BNC corpus \cite{leech1992100} to compute the association measures. In Table
\ref{tab:S10} and \ref{tab:S15} we report the average results on the S10 and S15
datasets, respectively. From these tables it is possible to see that the
performances of the system are highly influenced by the combination of measures
used.

\begin{table}
\begin{center}
  {\setlength{\extrarowheight}{2pt}%
\begin{tabular}{l c c c c c c c c} \hline
& dice & mdice & pmi & t-score & z-score & odds-r & chi-s & chi-s-c \\ \hline
$tfidf$ & 55.5 & 56.3 & 50.6 & 45.4 & 50.1 & 49.8 & 39.1 & 54.4 \\
$tfidf_{ext}$ & \textbf{56.5} & 55.9 & 50.1 & 45.0 & 49.9 & 49.5 & 39.1 & 54.2  \\
$vec$ & 54.7 & 54.3 & 49.3 & 44.1 & 49.4 & 53.6 & 39.3 & 50.5  \\
$vec_{ext}$ & 55.0 & 54.3 & 48.8 & 43.8 & 48.6 & 53.6 & 39.1 & 49.9  \\
$jcn$ & 51.3 & 50.6 & 40.1 & 50.1 & 47.6 & 52.6* & 50.1 & 50.6  \\
$wup$ & 37.2 & 36.9 & 35.6 & 32.2 & 37.9 & 36.8 & 38.4 & 35.4  \\
\end{tabular}}
\end{center}
\caption{Results as F1 for S10. The first result with a statistically significant difference from the best (bold result) is marked with * ($\chi^2$, $ p < 0.05 $).}
\label{tab:S10} 
\end{table} 
\begin{table}
\begin{center}
  {\setlength{\extrarowheight}{2pt}%
\begin{tabular}{l c c c c c c c c} \hline
& dice & mdice & pmi & t-score & z-score & odds-r & chi-s & chi-s-c \\ \hline
$tfidf$ & 64.1 & 64.2 & 63.1 & 59.0 & 61.8 & \textbf{65.3} & 63.3* & 62.4 \\
$tfidf_{ext}$ & 62.9 & 63.1 & 62.4 & 58.7 & 60.9 & 63.0 & 62.0 & 61.1 \\
$vec$ & 62.8 & 62.3 & 62.8 & 59.8 & 62.3 & 62.9 & 61.1 & 60.3 \\
$vec_{ext}$ & 60.5 & 59.9 & 61.2 & 57.8 & 59.7 & 60.6 & 60.1 & 59.4 \\
$jcn$ & 57.2 & 57.6 & 56.7 & 57.9 & 57.0 & 56.9 & 57.5 & 57.6 \\
$wup$ & 46.2 & 45.4 & 43.8 & 45.4 & 45.9 & 47.4 & 46.1 & 45.5 \\
\end{tabular}}
\end{center}
\caption{Results as F1 for S15. The first result with a statistically significant difference from the best (bold result) is marked with * ($\chi^2$, $ p < 0.05 $).}
\label{tab:S15} 
\end{table} 

\begin{figure} \includegraphics[width=1\textwidth]{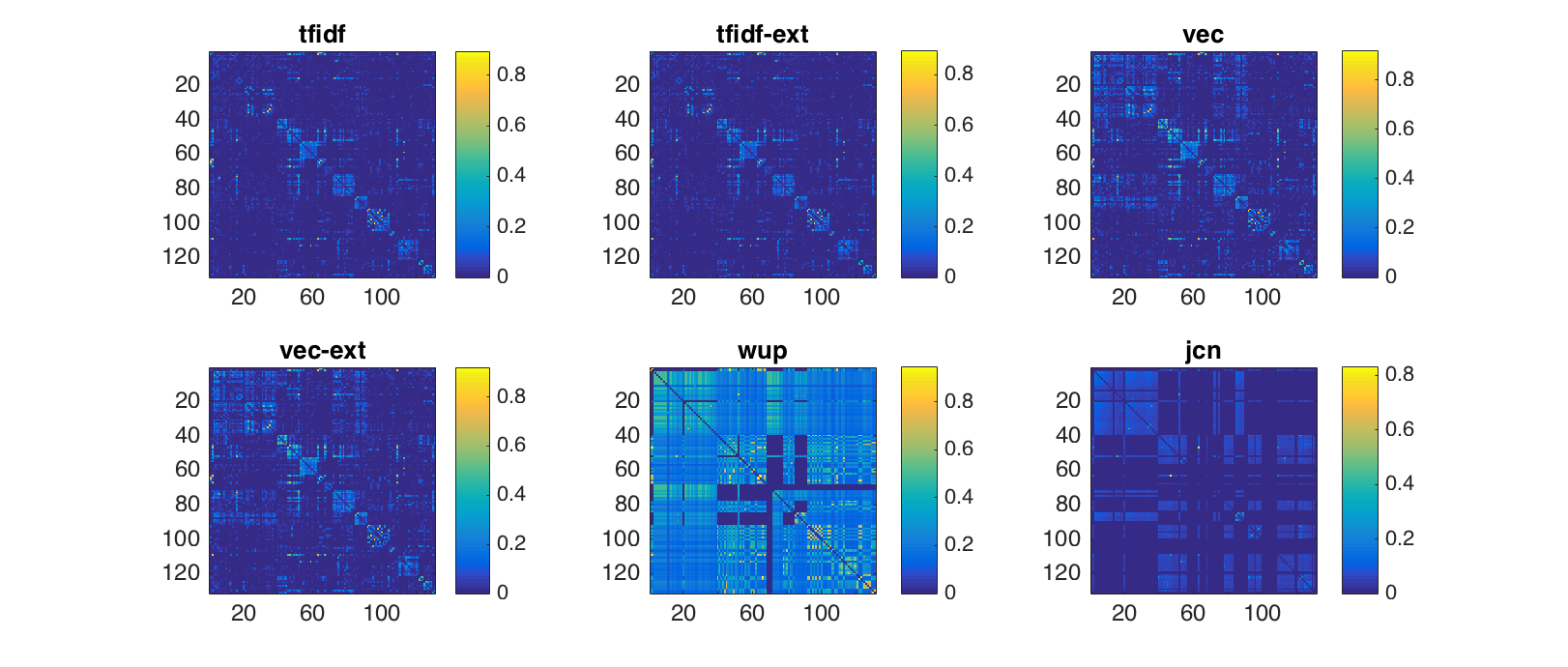} 
  \caption{The representations of the payoff matrix $Z$ computed on three
sentences of the second text of S10, with the measures described in Section
\ref{sec:semMes}. All the senses of the words in the text are sequentially ordered.}
   \label{fig:plotWsim} \end{figure}

\begin{figure} \includegraphics[width=1\textwidth]{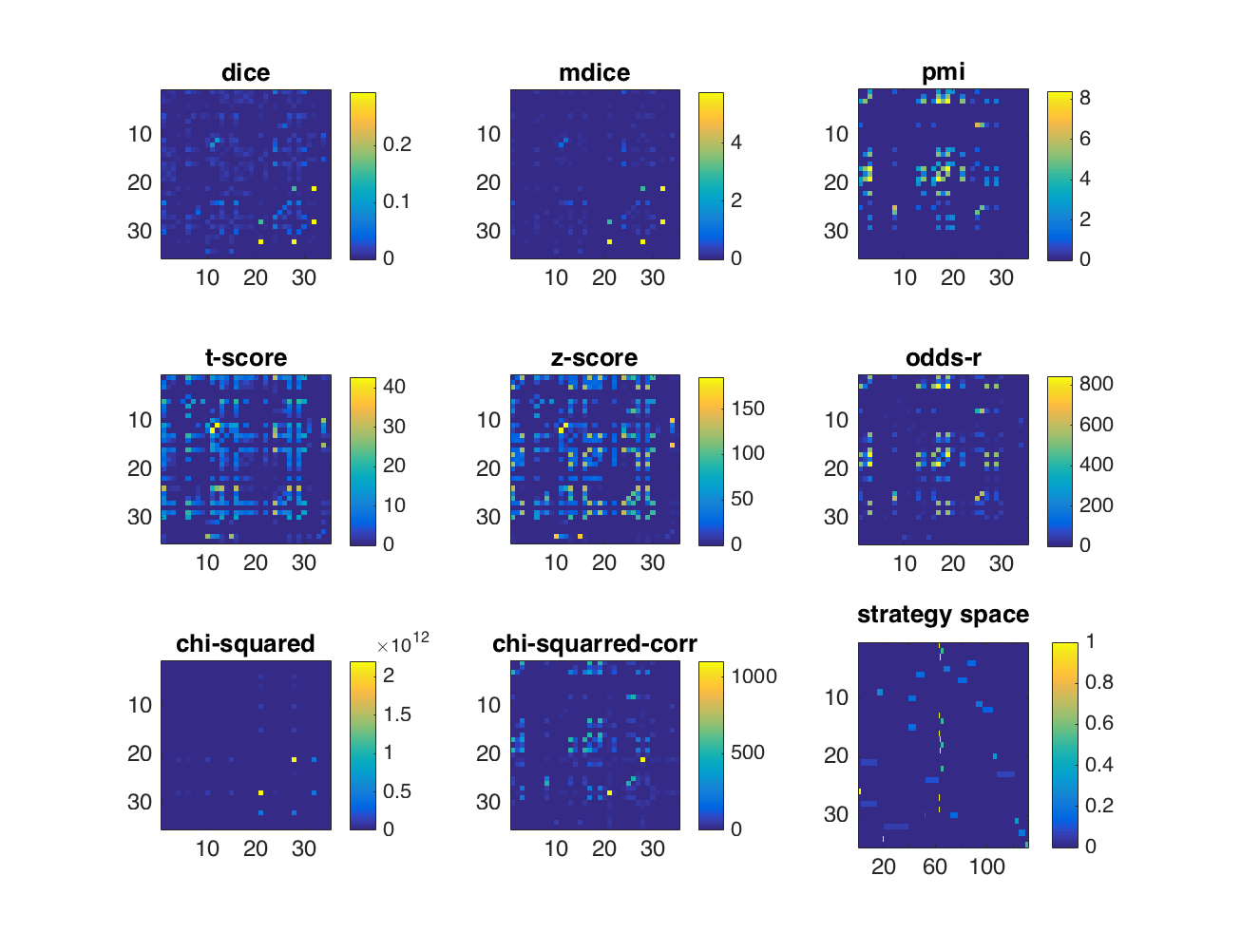}
\caption{The representations of the adjacency matrix of the graph $W$ computed
on three sentences of the second text of S10, with the measures described in
Section \ref{sec:disMes}. The words are ordered sequentially and reflect the
list proposed above. For a better visual comparison only positive values are
presented, whereas the experiments are performed considering also negative
values. The last image represents the strategy space of the players.}
\label{fig:plotW} \end{figure}

As an example of the different representations generated by the measures
described in Section \ref{sec:impDetails} we can observe Figure
\ref{fig:plotWsim} and \ref{fig:plotW}, which depict the matrices $Z$ and the adjacency matrix of the graph $W$,
respectively and are computed on the following three sentences from the second
text of S10,
\\

\begin{minipage}{0.85\textwidth} \leftskip1em \emph{The rivers Trent and Ouse, which provide the main fresh
water flow into the Humber, drain large industrial and urban areas to the
south and west (River Trent), and less densely populated agricultural areas
to the north and west (River Ouse). The Trent/Ouse confluence is known as Trent
Falls. On the north bank of the Humber estuary the principal river is the
river Hull, which flows through the city of Kingston-upon-Hull, and has a
tidal length of 32 km, up to the Hempholme Weir.} \vspace{1em} \end{minipage}

\noindent resulting in 35 content words (names and verbs) listed below and 131 senses.%
\begin{multicols}{4}\begin{enumerate}\footnotesize{
\item river~n
\item Trent~n
\item Ouse~n
\item provide~v
\item main~n
\item water~n
\item flow~n
\item Humber~n
\item drain~v
\item area~n
\item south~n
\item west~n
\item River~n
\item Trent~n
\item area~n
\item River~n
\item Ouse~n
\item Trent~n
\item Ouse~n
\item confluence~n
\item be~v
\item Trent~n
\item Falls~n
\item bank~n
\item Humber~n
\item estuary~n
\item river~n
\item be~v
\item river~n
\item flow~v
\item city~n
\item have~v
\item length*~n
\item km~n
\item Weir~n}
\end{enumerate}
\end{multicols}%
The first observation that can be done on the results is related to the semantic
measures; in fact, the relatedness measures perform significantly better than
the semantic similarity measures. This is due to the fact that \emph{wup} and
\emph{jcn} can be computed only on synsets, which have the same part of speech.
This limitation affects the results of the algorithm because the games played
between two words with different parts of speech have no effect on the dynamics
of the system, since the values of the resulting payoff matrices are all zeros.
This affects the performances of the system in terms of recall, because, in this
situation, these words tend to remain on the central point of the simplex and
also in terms of precision, because the choice of the meaning of a word is
computed only taking into account the influence of words with the same part of
speech. In fact, from Figure \ref{fig:plotWsim} we can see that the
representations provided by \emph{wup} and \emph{jcn}, for the text described
above, have many uniform areas, this means that these approaches are not able to
provide a clear representation of the data. To the contrary, the representations
provided by the relatedness measures show a block structure on the main diagonal
of the matrix, which is exactly what is required for a similarity measure. The
use of the \emph{tf-idf} weighting schema seems to be able to reduce the noise in the
data representation, in fact the weights on the left part of the matrix are
reduced by \emph{tfidf} and \emph{tfidf-ext} whereas they have high values in
\emph{vec} and \emph{vec-ext}. The representations obtained with eXtended
WordNet are very similar to those obtained with WordNet and also their
performances are very close, although on average WordNet outperform eXtended
WordNet.

If we observe the performances of the association measures we can notice that on
average the best measures are \emph{dice}, \emph{mdice}, \emph{chi-s-c} and
also \emph{odds-r} on S15, to the contrary the other measures perform almost
always under the statistical significance. Observing the representations in
Figure \ref{fig:plotW} we can see that \emph{dice} and \emph{mdice} have a
similar structure, the difference between these two measures are that
\emph{mdice} has values on a different range and tends to differentiate better
the weights, whereas in \emph{dice} the values are almost uniform. \emph{Pmi}
tends to take high values when one word in the collocation has low frequency but
this does not imply high dependency and therefore compromise the results of the
games. From its representation we can observe that its structure is different
from the previous two, in fact it concentrate its values on collocations such as
\emph{river Trent} and \emph{river Ouse} and this has the effect to unbalance
the data representation. In fact the \emph{dice} and \emph{mdice} concentrate
their values on collocations such as \emph{river flow} and \emph{bank estuary}.
\emph{T-score} and \emph{z-score} have a similar structure, the only
difference is in the range of the values. For these measures we can see that the
distribution of the values is quite homogeneous meaning that these measures are
not able to balance well the weights. On \emph{odds-r} we can recognize a
structure similar to that of \emph{pmi}, the main difference is that it works on
a different range. The values obtained with \emph{chi-s} are on a wide range,
which compromises the data representation; in fact its results are always under
the statistical significance. \emph{Chi-s-c} works on a narrower range than
\emph{chi-s} and its structure resample that of \emph{dice}, in fact its results
are often high.

\subsubsection{N-gram Graph}\label{sec:parTunK} The association measures are able
to give a good representation of the text but in many cases it is possible that
a word in a specific text is not present in the corpus on which these measures
are calculated, furthermore, it is possible that these words are used with
different lexicalizations. A way to overcome these problems is to increase the
values of the nodes near a determined word, in this way it is possible to ensure
that the nodes in $W$ are always connected. Furthermore, it allows to exploit
local information, increasing the importance of the words, which share a
proximity relation with a determined word, in this way it is possible to give
more importance to (possibly syntactically) related words, as described in
Section \ref{sec:grCns}.

\begin{figure}[ht!] \begin{center}
\subfigure[S10]{
\label{fig:S10ktuning}
\includegraphics[width=0.5\textwidth]{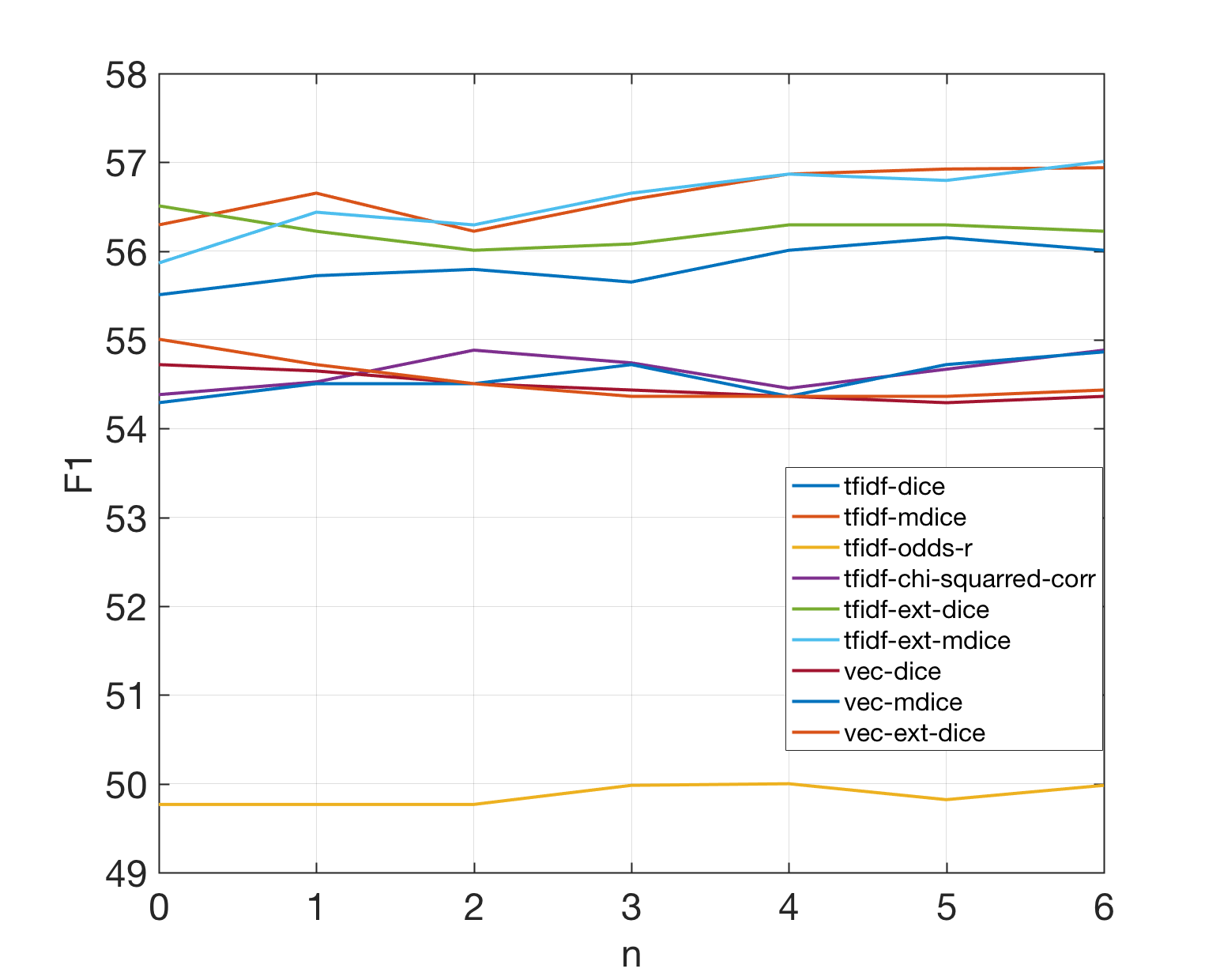} 
}
\subfigure[S15]{
\label{fig:S15ktuning}
\includegraphics[width=0.5\textwidth]{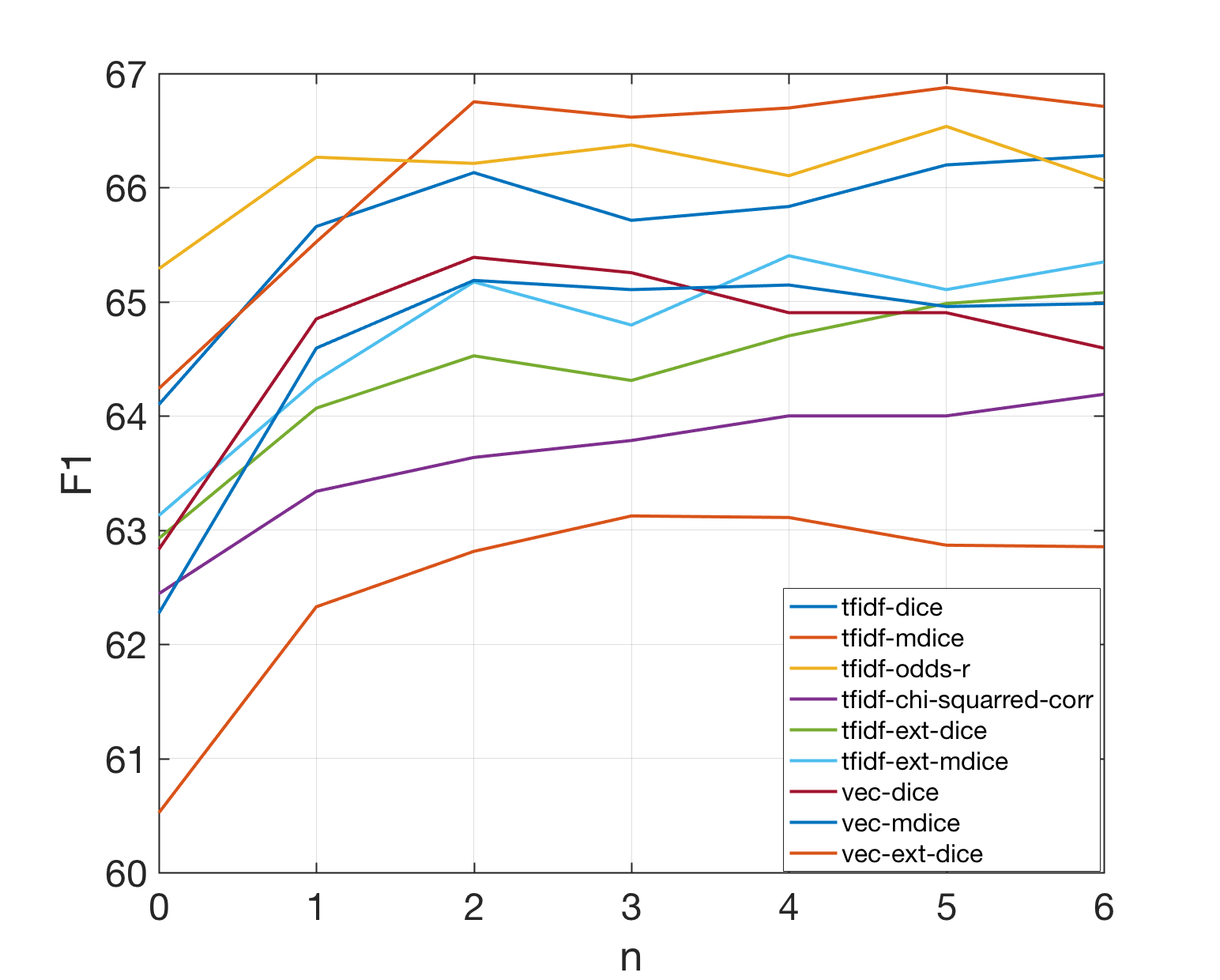}
}
\end{center}%
\caption{Results as F1 on S10 (on the left) and S15 (on the right) with
increasing values of neighbor nodes ($n$).}
\label{fig:Ktuning} \end{figure}

To test the influence that the parameter of the n-gram graph has on the
performances of the algorithm we selected the association and relatedness
measures with the highest results and conduct a series of experiments on the
same datasets presented above, with increasing values of $n$. The results of
these experiments on S10 and S15 are presented in Figure \ref{fig:S10ktuning}
and \ref{fig:S15ktuning}, respectively. From the plots we can see that this
approach is always beneficial for S15 and that the results increased
substantially with values of $n$ greater than 2. To the contrary on S10 this
approach is not always beneficial but in many cases it is possible to notice an
improvement. In particular we can notice that the pair of measures with highest
results on both datasets is \emph{tfidf-mdice} with $n=5$. This confirms also our earlier
experiments in which we have seen that these two measures are particularly
suited for our algorithm.
\subsubsection{Geometric Distribution}\label{sec:parTunP}%
Once we have identified the measures to use in our unsupervised system we can
test what is the best parameter to use in case we want to exploit information from sense labeled corpora. To tune
the parameter of the geometric distribution (described in Section
\ref{sec:strSpImp}) we used the pair of measures and the value of $n$
detected with the previous experiments and ran the algorithm on S10 and S15
with increasing values of $p$, in the interval $[0.05,0.95]$.

\begin{figure}[ht!] \begin{center}
\subfigure[Tuning.]{%
\label{fig:pTuning}
\includegraphics[width=0.5\textwidth]{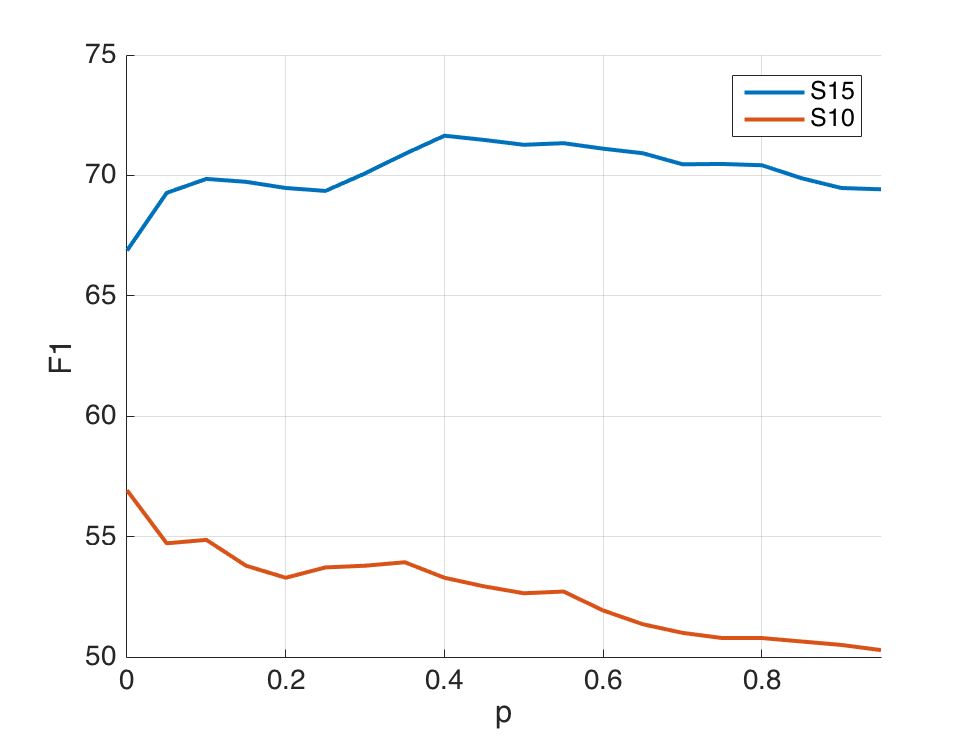} 
}%
\subfigure[Distributions.]{%
\label{fig:geoDist}
\includegraphics[width=0.5\textwidth]{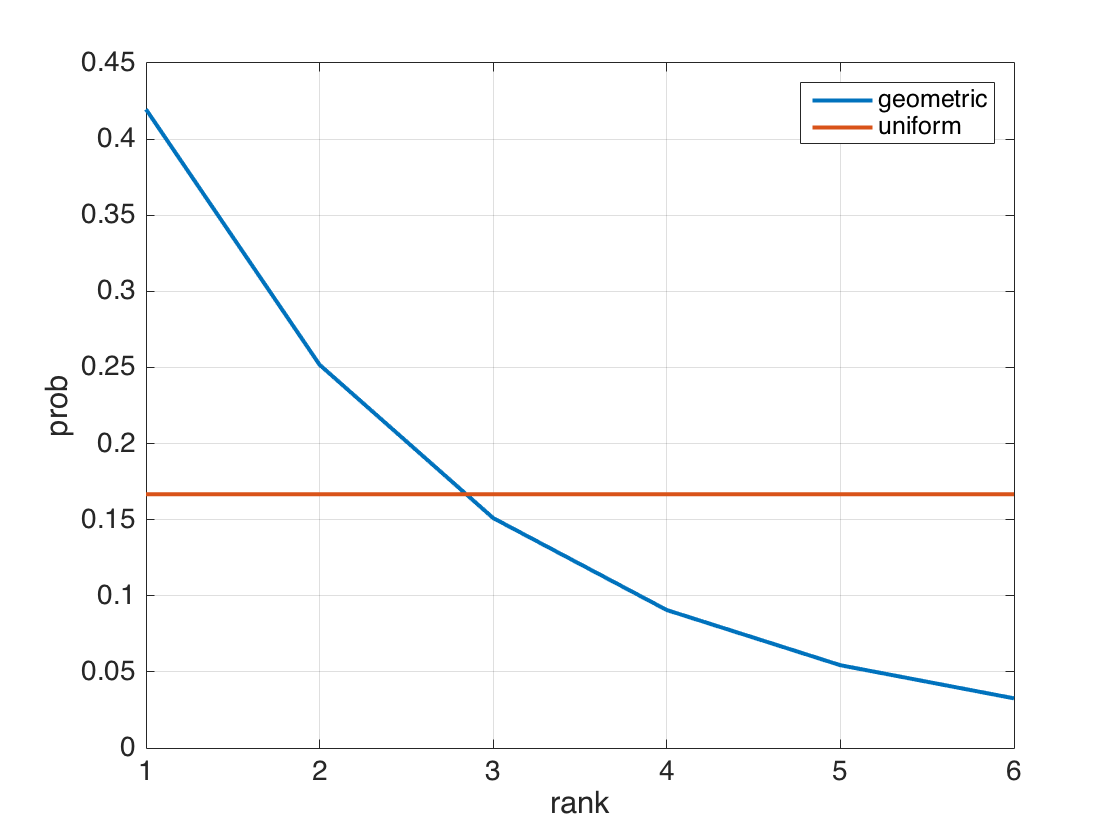}%
}%
\end{center}%
\caption{Results as F1 on S10 and S15 with increasing values of $p$ (on the
left), $p=0$ corresponds to the results with the unsupervised setting (on the left). An
example of geometric distribution with 6 ranked senses compared to the uniform distribution (on the right). }%
\label{fig:Ptuning} \end{figure}%
%
%
%
The results of this experiment are presented in Figure \ref{fig:pTuning}, where
we can see that the performances of the semi-supervised system on S15 are always
better those obtained with the unsupervised system ($p=0$). To the contrary, the
performances on S10 are always lower than those obtained with the unsupervised
system. This behavior is not surprising because the target words of S10 belong
to a specific semantic domain. We used SemCor to obtain the information about
the sense distributions and this resource is a general domain corpus, which is
not tailored for this specific task. In fact, as pointed out by McCarthy et. al
\shortcite{mccarthy2007unsupervised} the distribution of word senses on specific
domains is highly skewed and for this reason the most frequent sense heuristic
calculated on general domains corpora, such as SemCor, is not beneficial for
this kind of texts.

From the plot we can see that on S15 the highest results
are obtained with values of $p$ ranging from $0.4$ to $0.7$ and for the
evaluation of our model we decided to use $p = 0.4$ as parameter for the
geometric distribution, since with this value we obtained the highest result.%
%
%
%
\subsubsection{Error Analysis}\label{sec:errorTuning} The main problems that we noticed analyzing the results of
previous experiments are related to the
semantic measures. As we pointed out in Section
\ref{sec:parTunMeas}, these measures can be computed only on synsets
with the same part of speech and this influences the results in terms of recall.
The adverbs and adjectives are not disambiguated
with these measures, because of the lack of payoffs. This does not happen only in case of function words with
low semantic content but also for verbs with a rich semantic content, such as
\emph{generate}, \emph{prevent} and \emph{obtain}. The use of the relatedness
measures reduces substantially the number of words that are not disambiguated.
With these measures a word is not disambiguated only in cases in which the concepts denoted by it are not covered enough by the reference corpus, for example in
our experiments we have that words such as \emph{drawn-out}, \emph{dribble} and
\emph{catchment} are not disambiguated.

To overcome this problem we have used the
n-gram graph to increase the weights among neighboring words. Experimentally we
noticed that when this approach is used with the relatedness measures it leads
to the disambiguation of all the target words and with $n\geq1$ we have
$precision = recall$. The use of this approach influences the results also in
terms of precision, in fact if we consider the performances of the system on the
word \emph{actor}, we pass from $F1 = 0$ ($n=0$) to $F1 = 71.4$ ($n=5$). This is
because the number of relations of the two senses (synsets) of the word \emph{actor} are
not balanced in WordNet 3.0, in fact \emph{actor} as \emph{theatrical performer} has 21
relations whereas \emph{actor} as \emph{person who acts and gets things done}
has only 8 relations and this can compromise the computation of
the semantic relatedness measures. It is possible to overcome this limitation
using the local information given by the n-gram graph, which allows to balance the
influence of words in the text.

\begin{figure} \includegraphics[width=1\textwidth]{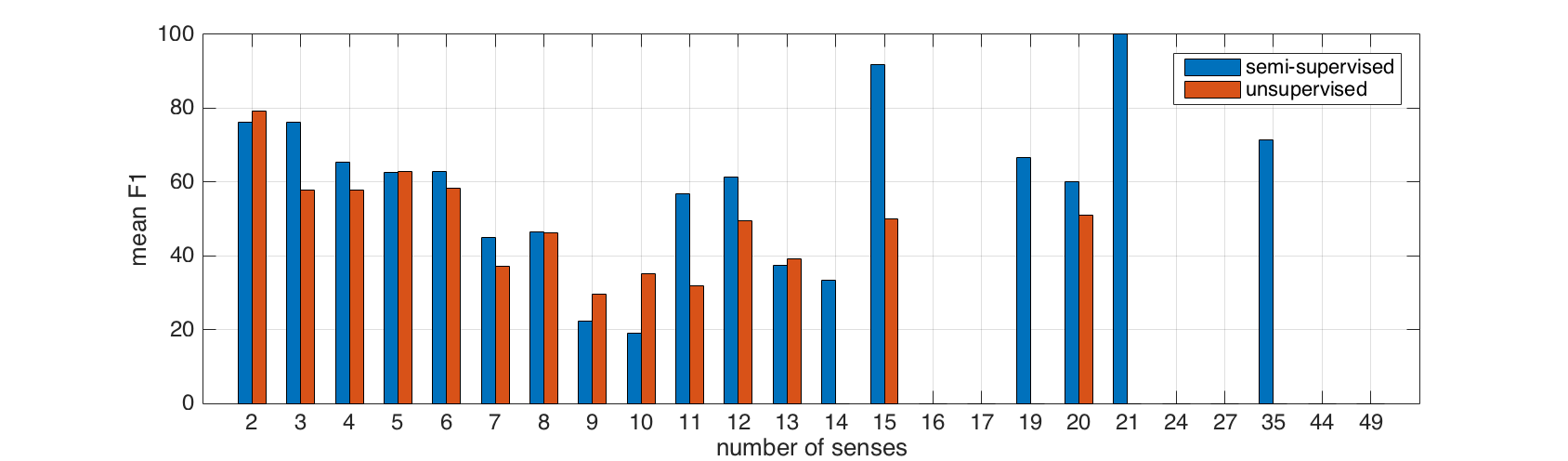}
\caption{Average F1 on the words of S15 grouped by number of senses, using the unsupervised and the semi-supervised system.}
\label{fig:errorSenses} \end{figure}

 Another aspect to consider is if the
polysemy of the words influences the results of the system. Analyzing the results
we noticed that the majority of the errors are made on words such as
\emph{make-v}, \emph{give-v}, \emph{play-v}, \emph{better-a}, \emph{work-v},
\emph{follow-v}, \emph{see-v}, \emph{come-v}, which have more that 20 different
senses and are very frequent words difficult to disambiguate in fine-grained
tasks. As we can see from Figure\ref{fig:errorSenses} this problem can be
partially solved using the semi-supervised system. In fact, the use of
information from sense-labeled corpora is particularly useful when the polysemy
of the words is particularly high.
\subsection{Evaluation Setup}\label{sec:evalSetup}%
We evaluated our algorithm with three fine-grained
datasets: Senseval-2 english all-words (S2) \cite{palmer2001english}, Senseval-3
english all-words (S3) \cite{snyder2004english}, SemEval-2007 all-words (S7)
\cite{pradhan2007semeval}, and one coarse-grained dataset, SemEval-2007 english
all-words (S7CG) \cite{navigli2007semeval}\footnote{We downloaded S2 from
www.hipposmond.com/senseval2, S3 from http://www.senseval.org/senseval3, S7 from
http://nlp.cs.swarthmore.edu/semeval/tasks/index.php and S7CG from
http://lcl.uniroma1.it/coarse-grained-aw}, using as knowledge base WordNet.
Furthermore we evaluated our approach on two datasets, SemEval-2013 task 12
(S13) \cite{navigli2013semeval} and KORE50 \cite{hoffart2012kore}\footnote{We
downloaded S13 from https://www.cs.york.ac.uk/semeval-2013/task12/index.html and
KORE50 from
http://www.mpi-inf.mpg.de/departments/databases-and-information-systems/research/yago-naga/aida/downloads/},
using as knowledge base BabeNet. 

We describe the
evaluation using as knowledge base WordNet in the next sections and in Section \ref{sec:evalBabel} we
present the evaluation conducted using as knowledge base BabelNet. We recall
that for all the next experiments we used \emph{mdice} to weight the
graph $W$, \emph{tfidf} to compute the payoffs, $n=5$ for the n-gram graph and
$p=0.4$ in case of semi-supervised learning. The results are provided as F1 for 
all the datasets except KORE50, for this dataset the results are provided as 
accuracy, as it is common in the literature.%

\subsubsection{Experiments Using WordNet as Knowledge Base}\label{sec:evalRisUns}
\begin{table} \begin{center} {\setlength{\extrarowheight}{1.5pt}
\begin{tabular}{ l c c c c c } \hline \multicolumn{6}{l}{SemEval 2007
coarse-grained - S7CG} \\ Method & All & N & V & A & R \\ \hline
WSD$_{games}^{uns}$ & 80.4 & 85.5 & 71.2 & 81.5 & 76.0 \\ 
WSD$_{games}^{ssup}$ & 82.8 & 85.4 & 77.2 & 82.9 & 84.6  \\ 
MFS & 76.3 & 76.0 & 70.1 & 82.0 & 86.0 \\ \hline \\ \hline
\multicolumn{6}{l}{SemEval 2007 fine-grained - S7} \\ 
Method & All & N & V & A & R \\ \hline
WSD$_{games}^{uns}$ & 43.3 & 49.7 & 39.9 &  $-$ &  $-$  \\ 
WSD$_{games}^{ssup}$ & 56.5 & 62.9 & 53.0 &  $-$ &  $-$ \\ 
MFS & 54.7 & 60.4 & 51.7 &  $-$ &  $-$ \\ \hline \\ \hline
\multicolumn{6}{l}{Senseval 3 fine-grained - S3} \\ 
Method & All & N & V & A & R \\ \hline
WSD$_{games}^{uns}$ & 59.1 & 63.3 & 50.7 & 64.5 & 71.4  \\ 
WSD$_{games}^{ssup}$ & 64.7 & 70.3 & 54.1 & 70.7 & 85.7   \\ 
MFS & 62.8 & 69.3 & 51.4 & 68.2 & 100.0 \\ \hline \\ \hline
\multicolumn{6}{l}{Senseval 2 fine-grained - S2} \\ 
Method & All & N & V & A & R \\ \hline
WSD$_{games}^{uns}$ & 61.2 & 69.8 & 41.7 & 61.9 & 65.1 \\ 
WSD$_{games}^{ssup}$ & 66.0 & 72.4 & 43.5 & 71.8 & 75.7 \\
MFS & 65.6 & 72.1 & 42.4 & 71.6 & 76.1 \\ \hline\end{tabular}} \end{center} 
\caption{Detailed results as F1 for the four datasets studied with \emph{tf-idf}
and $mdice$ as measures. The results show the performances of our unsupervised
(\emph{uns}) and semi-supervised (\emph{ssup}) system and the results obtained
employing the most frequent sense heuristic (MFS). Detailed information about
the performances of the systems on different part of speech are provided: nouns
(N), verbs (V), adjectives (A), adverbs (R). } \label{tab:detailedRis} 
\end{table}%
Table \ref{tab:detailedRis} shows the results as F1 for the four datasets that
we used for the experiments with WordNet. The table includes the results for the
two implementations of our system: the unsupervised and the semi-supervised and
the results obtained using the most frequent sense heuristic. For the
computation of the most frequent sense we assigned to each word to be
disambiguated the first sense returned by the WordNet reader provided by the
Natural Language Toolkit (version 3.0) \cite{bird2006nltk}. As we can see the
best performances of our system are obtained on nouns, on all the datasets. This
is in line with state of the art systems because in general the nouns have lower
polysemy and higher inter-annotator agreement \cite{palmer2001english}.
Furthermore, our method is particularly suited for nouns. In fact, the
disambiguation of nouns benefits from a wide context and local collocations
\cite{agirre2007word}. 

We obtained low results on verbs, on all datasets.
This, as pointed out by Dang \shortcite{dang2004investigations}, is a common
problem not only for supervised and unsupervised WSD systems but also for humans
which in many cases disagree about what constitutes a different sense for a
polysemous verb, compromising the sense tagging procedure. 

As we have
anticipated in Section \ref{sec:parTunP}, the use of prior knowledge is
beneficial for this kind of dataset. As we can see in Table
\ref{tab:detailedRis} using a semi-supervised setting improves the results of
5\% on S2 and S3 and of 12\% on S7. The big improvement obtained on S7 can be
explained by the fact that the results of the unsupervised system are well below
the most frequent sense heuristic, so exploiting the evidence from sense-labeled
dataset is beneficial. For the same reason, the results obtained on S7CG with a
semi-supervised setting are less impressive than those obtained with the
unsupervised systems; in fact, the structure of the datasets is different
and the results obtained with the unsupervised setting are well
above the most frequent sense. These series of experiments confirm that the use 
of prior knowledge is beneficial in general domain datasets and that when it is 
used the system performs better than the most common sense heuristic computed on the same corpus.

agraph{Comparison to State-of-the-Art Algorithms}\label{sec:evalSota}
\begin{table}
\begin{center}
\begin{tabular}{l l c c c c c } \midrule
& & S7CG & S7CG (N) & S7 & S3 & S2  \\ \midrule
\multirow{3}{*}{\begin{turn}{90}\footnotesize{unsup.}\end{turn}}& \emph{Nav10} &  $-$ &  $-$ & 43.1 & 52.9 &  $-$ \\
& \emph{PP$R_{w2w}$} & 80.1 & 83.6 & 41.7 & 57.9 & 59.7 \\
& \emph{WS$D_{games}$} & \textbf{80.4*} & \textbf{85.5} & \textbf{43.3} & \textbf{59.1} & \textbf{61.2} \\ \midrule
\multirow{6}{*}{\begin{turn}{90}\footnotesize{semi sup.}\end{turn}}& \emph{IRST-DDD-00} &  $-$ &  $-$ &  $-$ & 58.3 &  $-$ \\
& \emph{MFS} & 76.3 & 77.4 & 54.7 & 62.8 & 65.6*  \\ 
& \emph{MRF-LP} &  $-$  &  $-$ & 50.6* & 58.6 & 60.5 \\ 
& \emph{Nav05} & \textbf{83.2} & 84.1 &  $-$ & 60.4 &  $-$ \\
& \emph{PP$R_{w2w}$} & 81.4 & 82.1 & 48.6 & 63.0 & 62.6 \\
& \emph{WS$D_{games}$} & 82.8 & \textbf{85.4} & \textbf{56.5} & \textbf{64.7*} & \textbf{66.0} \\ \midrule
\multirow{2}{*}{\begin{turn}{90}\footnotesize{sup.}\end{turn}}&\emph{Best} & 82.5 & 82.3* & \textbf{59.1} & 65.2 & \textbf{68.6} \\
& \emph{Zhong10} & \textbf{82.6} & $-$ & 58.3 & \textbf{67.6} & 68.2 \\ \end{tabular}
\end{center} \caption{Comparison with state-of-the-art algorithms: unsupervised
(\emph{unsup.}), semisupervised (\emph{semi sup.}) and supervised (\emph{sup.}).
\emph{MFS} refers to the MFS heuristic computed on SemCor on each dataset and
\emph{BEST} refers to the best supervised system for each competition. The
results are provided as F1 and the first result with a statistically significant
difference from the best of each dataset is marked with * ($\chi^2$, $ p < 0.05
$).}
\label{tab:sotaComp}
\end{table}
Table \ref{tab:sotaComp} shows the results of our system and the results
obtained by state-of-the-art systems on the same datasets. We compared our
method with supervised, unsupervised and semi-supervised system on four
datasets. The supervised systems are \emph{It makes sense} \cite{zhong2010makes}
(\emph{Zhong10}), an open source WSD system based on support vector machines
\cite{steinwart2008support}; and the best system that participated to each
competition (\emph{Best}). The semi-supervised systems are: \emph{IRST-DDD-00}
\cite{strapparava2004pattern}, based on WordNet domains and on manually
annotated domain corpora; \emph{MFS}, which corresponds to the most frequent
sense heuristic implemented using the WordNet corpus reader of the natural
language toolkit; \emph{MRF-LP} based on Markov random field
\cite{chaplot2015unsupervised}; \emph{Nav05} \cite{navigli2005structural} a
knowledge based method that exploits manually disambiguated word senses to
enrich the knowledge base relations; $PPR_{w2w}$ \cite{agirre2014random} a
random walk method that uses contextual information and prior knowledge about
senses distribution to compute the most important sense in a network given a
specific word and its context. The unsupervised systems are: \emph{Nav10}, a
graph based WSD algorithm that exploits connectivity measures to determine the
most important node in the graph composed by all the senses of the words in a
sentence; and a version of the $PPR_{w2w}$ algorithm that does not use sense
tagged resources.

The results show that our unsupervised system performs
better than any other unsupervised algorithm in all datasets. In S7CG and S7 the
difference is minimal compared with $PPR_{w2w}$ and \emph{Nav10}, respectively;
in \emph{S3} and \emph{S2} the difference is more substantial compared to both
unsupervised systems. Furthermore, the performances of our system is more stable
on the four datasets, showing a constant improvement on the state-of-the-art.

 The comparison with semi supervised systems shows that our system performs
always better than the most frequent sense heuristic when we use information
from sense-labeled corpora. We can note a strange behavior on S7CG, when we use
prior knowledge the performances of our semi-supervised system are lower than our
unsupervised system and state-of-the-art. This is because on this dataset the
performances of our unsupervised system are better than the results than can be
achieved by using labeled data to initialize the strategy space of the semi
supervised system. On the other three datasets we can note a substantial
improvement in the performances of our system, with stable results higher than
state-of-the-art systems. 

 Finally we can note that the results of our semi
supervised system, on the fine-grained datasets, are close to the performances of
state-of-the-art supervised systems, with values that are statistically
relevant only on $S3$. We can also note that the performances of our system on
the nouns of the S7CG dataset are higher than the results of the supervised
systems.
\subsubsection{Experiments with BabelNet}\label{sec:evalBabel} BabelNet is
particularly useful when the number of named entities to disambiguate is high.
In fact it is not possible to perform this task using only WordNet, because its
coverage on named entities is limited. For the experiments on this section we
used BabelNet to collect the sense inventories of each word to be disambiguated,
the \emph{mdice} measure to weight the graph $W$ and NASARI to obtain the
semantic representation of each sense. The similarity among the representation
obtained with this resource are computed using the cosine similarity measure,
described in Section \ref{sec:semMes}. The only differences with the experiments
presented in Section \ref{sec:evalRisUns} are that we used BabelNet as knowledge
base and NASARI as resource to collect the sense representations instead of
WordNet.

 S13 consists of 13 documents in different domains, available in 5
languages (we used only English). All the nouns in these texts are annotated
using BabelNet, with a total number of 1931 words to be disambiguated (English
dataset). KORE50 consists of 50 short English sentences with a total number of
146 mentions manually annotated using YAGO2 \cite{hoffart2013yago2}. We used the
mapping between YAGO2 and Wikipedia to obtain for each mention the corresponding
BabelNet concept, since there exists a mapping between Wikipedia and BabelNet.
This dataset contains highly ambiguous mentions, which are difficult to capture
without the use of a large and well organized knowledge base. In fact, the
mentions are not explicit and require the use of common knowledge to identify
their intended meaning. 

 We used the SPARQL
endpoint\footnote{http://babelnet.org/sparql/} provided by BabelNet to collect
the sense inventories of the words in the texts of each dataset. For this task
we filtered the first 100 resources whose label contains the lexicalization of
to word to be disambiguated. This operation can increase the dimensionality of
the strategy space, but it is required because especially in KORE50 there are many
indirect references, such as Tiger to refer to Tiger Woods (the famous golf
player) or Jones to refer to John Paul Jones (the Led Zeppelin bassist).

agraph{Comparison to State-of-the-Art Algorithms}\label{sec:evalSotaBabel}
\begin{table} \begin{center}
\begin{tabular}{l c c } \hline & S13 & KORE50 \\ \hline \emph{WS$D_{games}$} &
\textbf{70.8} & \textbf{75.7} \\ \emph{Babelfy} & 69.2 & 71.5 \\ \emph{SUDOKU} &
66.3 &  $-$ \\ \emph{MFS} & 66.5* &  $-$ \\ \emph{PP$R_{w2w}$} & 60.8 &  $-$ \\
\emph{KORE} &  $-$ & 63.9* \\ \emph{GETALP} & 58.3 &  $-$ \\ \end{tabular} \end{center}
\caption{Comparison with state-of-the-art algorithms on WSD and entity linking.
The results are provided as F1 for S13 and as accuracy for KORE50. The first
result with a statistically significant difference from the best (bold result)
is marked with * ($\chi^2$, $ p < 0.05 $).} \label{tab:sotaCompBabel}
\end{table}%
The results of these experiments are shown in Table \ref{tab:sotaCompBabel},
where it is possible to see that the performances of our system are close to the
results obtained with Babelfy on S13 and substantially higher on KORE50. This is
because with our approach it is necessary to respect the textual coherence,
which is required when a sentence contains a high level of ambiguity, such as
those proposed by KORE50. To the contrary, PP$R_{w2w}$ performs poorly on this
dataset. This because, as attested in \cite{moro2014entity}, it disambiguates
the words independently, without imposing any consistency requirements. 

 The
good performances of our approach are also due to the good semantic
representations provided by NASARI, in fact, it is able to exploit a richer
source of information, Wikipedia, which provides a larger coverage and a wider
source of information than WordNet alone.

The results on KORE50 are presented as accuracy, following the custom of
previous work on this dataset. As we have anticipated it contains
decontextualized sentences, which require common knowledge to be disambiguated.
This common knowledge is obtained exploiting the relations in BabelNet that
connect related entities but in many cases this is not enough because the
references to entities are too general and in this case the system is not able
to provide an answer. It is also difficult to exploit distributional
information on this dataset, since the sentences are short and in many cases
cryptic. For these reasons the recall on this dataset is well below the
precision: $55.5\%$. The system does not provide answers for the entities in
sentences such as: \emph{Jobs and Baez dated in the late 1970s, and she
performed at his Stanford memorial}, but it is able to disambiguate correctly 
the same entities in sentences where there is more contextual information.
\section{Conclusions} In this article we have introduced a new method for WSD
based on game theory. We have provided an extensive introduction on the WSD task
and explained the motivations behind the choice to model the WSD problem as a
constraint satisfaction problem. We conducted an extensive series of
experiments to find out the similarity measures that perform better in our
framework. We have also evaluated our system with two different implementations
and compared our results with state-of-the-art systems, on different WSD tasks.

 Our method can be considered as a continuation of knowledge based, graph
based and similarity based approaches. We used the methodologies of these three
approaches combined in a game theoretic framework. This model is used to perform
a consistent labeling of senses. In our model we try to maximize the textual
coherence imposing that the meaning of each word in a text must be related to
the meaning of the other words in the text. To do this we exploited
distributional and proximity information to weight the influence that each word
has on the others. We exploited also semantic similarity information to weight
the strengths of compatibility among two senses. This is of great importance
because it imposes constraints on the labeling process, developing a contextual
coherence on the assignment of senses. The application of a game theoretic
framework guarantees that these assumptions are met. Furthermore, the use of the
replicator dynamics equation allows to always find the best labeling assignment.

 Our system in addition to have a solid mathematical and linguistic
foundation, has demonstrated to perform well compared with state-of-the-art
systems and to be extremely flexible. In fact, it is possible to implement new
similarity measures, graph constructions and strategy space initializations to
test it in different scenarios. It is also possible to use it as completely
unsupervised or to use information from sense-labeled corpora. 

 The features
that make our system competitive, compared with state-of-the-art systems, are
that instead of finding the most important sense in a network to be associated
to the meaning of a single word, our system disambiguates all the words at the
same time taking into account the influence that each word has on the others and
imposes to respect the sense compatibility among each sense before to assign a
meaning. We have demonstrated how our system can deal with sense shifts, where a
centrality algorithm, which tries to find the most important sense in a network
can be deceived by the context. In our case, the weighting of the context ensures to
respect the proximity structure of a sentence and to
disambiguate each word according to the context in which it appears. This is
because the meaning of a word in a sentence does not depend on \emph{all} the
words contained in the sentence but only on those that share a proximity (or syntactical)
relation and those with which enjoy a high distributional similarity.

\begin{acknowledgments} This work was supported by Samsung Global Research
Outreach Program. We are deeply grateful to Rodolfo Delmonte for his insights on
the preliminary phase of this work and to Bernadette Sharp for her help during
the final part of it. We would also like to thanks Phil Edmonds for providing us
the correct version of the Senseval 2 dataset. \end{acknowledgments} 

\bibliographystyle{fullname}
\bibliography{bibliography}

\end{document}